\documentclass[preprint,12pt,3p]{elsarticle}

\usepackage{amssymb}

\usepackage{algorithm}
\usepackage{algorithmic}

\usepackage{amsmath}

\usepackage{array,multirow,graphicx} 
\usepackage{colortbl}
\usepackage[table]{xcolor} 

\usepackage{setspace}

\spacing{1.5}

\journal{Biomedical Informatics}

\begin{document}

\begin{frontmatter}

\title{Benchmarking Relief-Based Feature Selection Methods for Bioinformatics Data Mining}

\author[label1]{Ryan J. Urbanowicz\corref{cor1}}
\address[label1]{Institute for Biomedical Informatics, University of Pennsylvania, Philadelphia, PA 19104, USA}

\cortext[cor1]{Corresponding Author}

\ead{ryanurb@upenn.edu}

\author[label1]{Randal S. Olson}
\ead{olsonran@upenn.edu}

\author[label1]{Peter Schmitt}
\ead{pschmitt@upenn.edu}

\author[label2]{Melissa Meeker}
\address[label2]{Ursinus College, Collegeville, PA, 19426, USA }
\ead{memeeker@ursinus.edu}

\author[label1]{Jason H. Moore}
\ead{jhmoore@upenn.edu}

\begin{abstract}
Modern biomedical data mining requires feature selection methods that can (1) be applied to large scale feature spaces (e.g. `omics' data), (2) function in noisy problems, (3) detect complex patterns of association (e.g. gene-gene interactions), (4) be flexibly adapted to various problem domains and data types (e.g. genetic variants, gene expression, and clinical data) and (5) are computationally tractable. To that end, this work examines a set of filter-style feature selection algorithms inspired by the `Relief' algorithm, i.e. Relief-Based algorithms (RBAs). We implement and expand these RBAs in an open source framework called ReBATE (Relief-Based Algorithm Training Environment). We apply a comprehensive genetic simulation study comparing existing RBAs, a proposed RBA called MultiSURF, and other established feature selection methods, over a variety of problems. The results of this study (1) support the assertion that RBAs are particularly flexible, efficient, and powerful feature selection methods that differentiate relevant features having univariate, multivariate, epistatic, or heterogeneous associations, (2) confirm the efficacy of expansions for classification vs. regression, discrete vs. continuous features, missing data, multiple classes, or class imbalance, (3) identify previously unknown limitations of specific RBAs, and (4) suggest that while MultiSURF* performs best for explicitly identifying pure 2-way interactions, MultiSURF yields the most reliable feature selection performance across a wide range of problem types.
\end{abstract}


\end{frontmatter}


\section{Introduction} \label{intro}
Feature selection is often an essential task in biomedical data mining and modeling (i.e. induction), particularly in problems where the data is noisy, complex, and/or includes a very large feature space. Many feature selection strategies have been proposed over the years, generally falling into one of three categories: (1) filter methods, (2) wrapper methods, or (3) embedded methods \citep{saeys2007review,bolon2013review,chandrashekar2014survey,tang2014feature,jovic2015review,mlambosurvey,urbanowicz2017relief}. Feature selection methods have further been characterized based on whether selection relies on scores assigned to individual features or instead to a candidate subset of features \citep{yu2004efficient,bolon2013review}.

The present study focuses on the family of \emph{Relief-based} feature selection methods referred to here as Relief-Based Algorithms (RBAs) that can be characterized as \emph{individual evaluation filter methods}. In work that pairs with this research paper, \citet{urbanowicz2017relief} introduced and surveyed RBA methods, detailing why they are advantageous in contrast with other feature selection methods. To summarize here, RBAs retain the general benefits of filter-methods, i.e. they are relatively fast (with an asymptotic time complexity of $\mathcal{O}(instances^2\cdot features)$), and the selected features are induction algorithm independent. More importantly, RBAs are the only filter-methods known that have the ability to capture feature dependencies in predicting endpoint/outcome, i.e. feature interactions, or more specifically gene-gene interactions \citep{bolon2013review}. This unique ability has been attributed to Relief's use of `nearest neighbor instances' in calculating feature weights \citep{kononenko1997overcoming,kononenko2008non}. 

The Relief algorithm concept has also been shown to be extendable to many different data type concerns including classification vs. regression, discrete vs. continuous features, missing data, multiple classes, and class imbalance. Unfortunately many RBAs and associated implementations have yet to be extended to data types beyond clean binary classification problems with discrete features. This is important to their application to analyses involving gene expression, quantitative traits, or integrated `omics' data. 

RBAs are also advantageous because they output individual feature weights. These weights can be used both to flexibly set different criteria for defining a feature subset as well as be applied to feature weighting schemes, where for example feature weights probabilistically guide machine learning modeling downstream \citep{urbanowicz2012using}. One other notable aspect of RBAs is that they do not eliminate feature redundancies (i.e. feature correlations). This could be viewed either as an advantage or disadvantage based on the problem at hand. For problems that require the removal of feature weights, many effective methods have been developed to remove feature redundancy as reviewed by \citet{urbanowicz2017relief}. Since this `drawback' of not handling feature redundancy is undisputed but there are independent effective methods available to deal with redundancy when needed, we do not consider the topic further in this research. 

\subsection{Core Relief Algorithms}
The present study specifically focuses on what we will refer to as `core' RBAs, or algorithm variants designed to be run for a single iteration through the training data.  In contrast, a handful RBA extensions have also been proposed to improve performance in very large feature spaces by applying a core Relief algorithm iteratively, e.g. I-RELIEFF \citep{sun2006iterative}, TuRF \citep{moore2007tuning}, evaporative cooling ReliefF \citep{mckinney2007evaporative}, and iVLSReliefF \citep{eppstein2008very}, or applying it to a multitude of random feature subsets rather than the entire feature space to improve efficiency, i.e. VLSReliefF \citep{eppstein2008very}. The iterative extensions are much more computationally expensive and the success of the VLSReliefF and iVLSReliefF methods rely on additional run parameters that may require problem domain knowledge to be set optimally.  

Stepping back, we propose that there are two larger Relief algorithm research questions. First, what is the most effective core Relief algorithm? This question is asked given the expectation that the performance of any core method alone will deteriorate as the feature space becomes very large \citep{moore2007tuning,eppstein2008very}. Second, what is the most effective iterative Relief expansion for improving performance in very large feature spaces? This work focuses exclusively on the first question. We expect that by first identifying and adopting the most reliable core algorithm this will maximize the performance of any iterative expansion, since they each rely on core algorithm functionality. However, it will be important to revisit strategies for very large feature spaces in future work. For a complete review of RBA research with respect to core, iterative, efficiency, and data type handeling methodologies we refer readers to \citet{urbanowicz2017relief}.

\subsection{Bioinformatics}
The focus of this work on RBAs is related to our interest in the application domain of bioinformatics. Most of the RBAs analyzed in this paper have been developed and applied within the purview of genetic association problems. Such problems are commonly characterized as (1) noisy, (2) having varied or sometimes mixed data types (e.g. discrete and continuous features), and (3) including very large feature spaces \citep{moore2007tuning,greene2009spatially,greene2010informative,granizo2013multiple}.  Also, the detection of complex patterns of association between features and the endpoint is of particular interest in bioinformatics. In particular, this study considers two important complicating phenomena: epistasis, i.e. feature interactions (or gene-gene interactions in the context of bioinformatics) \citep{cordell2002epistasis,moore2005traversing} and heterogeneous associations with endpoint, i.e. genetic heterogeneity or phenocopy (in the context of bioinformatics problems) \citep{urbanowicz2010application,urbanowicz2013role}.  Genetic heterogeneity occurs when the same or similar phenotypic endpoint can be the result of distinct independent relevant features (or set of relevant features) in different subsets of the sample population. Heterogeneous associations are recognized to confound most modeling techniques \citep{ritchie2003power}, with the exception of induction algorithms that can distinguish problem `niches' such as rule-based machine learning approaches \citep{urbanowicz2015exstracs}. While not all problem domains may be as complex as those in bioinformatics, we expect the findings of this work to be applicable to any data mining problems calling for feature selection.

Existing RBAs examined in this study have shown promise, however there are few guidelines suggesting which RBAs, or feature selection methods in general, to apply to various bionformatics problems. Further a large scale comparison of core RBAs has not yet been achieved to answer the initial questions of which method is most likely to capture the greatest breadth of underlying associations in the dataset (i.e. main, interaction, heterogeneous effects), and how is performance of these various algorithms impacted by dataset characteristics expected across a breadth of bioinformatics applications (e.g. magnitude of noise, number of features, missing data, imbalanced data, continuous-valued features, quantitative traits, and a mix of discrete and continuous features)? 

\subsection{Study Overview}
In the present study, we (1) implement a variety of core RBAs as part of an accessible, open source Python software package called ReBATE, (2) introduce a new core RBA called MultiSURF, (3) extend all implemented algorithms to be able to accommodate varied data type issues, i.e. binary classification, multi-class classification, or regression, discrete, continuous or mixed feature types, missing data and class imbalance, (4) design, generate, and apply a comprehensive simulation study of 2280 datasets to validate the efficacy of our data type extensions, and to compare the efficacy of 13 feature selection methods, i.e. eight RBAs, three traditional `filter' feature selection methods, and two `wrapper' feature selection methods, (5) investigate the reasons for identified performance discrepancies among feature selection methods, (6) identify the best performing and most reliable feature selection algorithms evaluated, and (7) organize what we have learned from this investigation to guide future RBA application and development.

The remainder of this paper is organized as follows: Section \ref{methods} describes the methods, Section \ref{results} presents the results, Section \ref{discussion} offers discussion regarding why specific strengths and weaknesses were observed for respective methods. Lastly, Section \ref{conclusions} offers conclusions and directions for future work.

\section{Methods} \label{methods}
In this section, we begin by describing the ReBATE software including: (1) the previously defined RBAs that it implements, (2) our proposed MultiSURF algorithm, and (3) the adopted data type extension strategies. Next, we describe the evaluations including: (1) other filter and wrapper based feature selection methods compared, (2) design of the simulation study, and (3) a discussion of the evaluation metrics. 

\subsection{ReBATE}
To facilitate the accessibility of various RBAs and promote their ongoing development and application, we have implemented Relief-Based Algorithm Training Environment (ReBATE). With ReBATE, we seek to balance data type flexibility, run time efficiency, and ease of development in a Python package framework. At the time of writing, ReBATE has been implemented with five core RBAs: i.e. ReliefF \citep{kononenko1994estimating}, SURF \citep{greene2009spatially}, SURF* \citep{greene2010informative}, MultiSURF* \citep{granizo2013multiple}, and our proposed MultiSURF algorithm. These core RBAs were chosen for this study because (1) they were explicitly developed for and previously evaluated on noisy and epistatic problems, (2) accessible Python implementations were available for each, and (3) they represented a competitive diversity of core RBAs.  Additionally, we included the iterative TuRF algorithm \citep{moore2007tuning} in ReBATE. However, as an iterative approach, TuRF is not evaluated further in this study. 

These five core RBAs and TuRF were originally implemented in the open source Multifactor Dimensionality Reduction (MDR) \citep{ritchie2001multifactor} software package\footnote{http://sourceforge.net/projects/mdr}. These Java implementations are computationally efficient, but can only handle complete data (i.e. no missing values) with discrete features and a binary endpoint. Python 2.7 versions of these algorithms were more recently implemented and made available within the open source Extended Supervised Tracking and Classifying System (ExSTraCS)\footnote{https://github.com/ryanurbs/ExSTraCS\_2.0} \citep{urbanowicz2014extended,urbanowicz2015exstracs}. These were less computationally efficient, but extended each algorithm to handle different data types including continuous features, multi-class endpoints, regression, and missing data. The ReBATE implementations of these RBAs have restructured these ExSTraCS implementations for efficiency and modularity, while preserving the ability to handle different data types.  Notably, while these data type expansions had been previously proposed and implemented within the ExSTraCS framework, they had yet to be experimentally validated.  
We have implemented ReBATE both as a stand alone software package\footnote{https://github.com/EpistasisLab/scikit-rebate} as well as a scikit-learn \citep{scikit-learn} compatible format\footnote{https://github.com/EpistasisLab/ReBATE}. It was our goal to make data type flexible implementations of these algorithms available for real-world application, as well as encourage ongoing methodological development or expansion of the ReBATE algorithm repertoire.  

ReBATE includes a data pre-processing step that automatically identifies essential data type characteristics. Specifically this includes (1) distinguishing discrete from numerical features, (2) distinguishing a discrete from numerical endpoint, (3) identifying the min-max value range for numerical features or endpoint, (4) for discrete classes, determining the number of unique classes (i.e. binary or multi-class) as well as the number of instances having each class label, and (5) identifying the presence of missing data, with a standard identifier, e.g. `N/A'. This pre-processing automates the adaptation of each RBA to the relevant data types.

In the following subsections, we provide methodological summaries of these five core algorithms.  They were described in contrast to the larger family of RBAs by \citet{urbanowicz2017relief}. Next, in the remaining subsections, we detail how these core RBAs have been universally extended to handle specific data-type challenges. Our focus in this study is to demonstrate that our adopted strategies are functional, but do not seek to claim that they are optimal. Currently there is minimal empirical evidence in the literature to support the conclusion that that any particular RBA data type handeling strategy performs optimally. 

\subsubsection{ReliefF} \label{ReliefF}
The original Relief algorithm \citep{kira1992practical,kira1992feature} was quickly improved upon to yield the most widely known RBA to date, ReliefF \citep{kononenko1994estimating}. For clarity, we will begin with a complete algorithmic description of ReliefF. Other complete descriptions of Relief and ReliefF can be found in \citep{kononenko1996relieff,kononenko1997overcoming,robnik2003theoretical}. For simplicity ReliefF is described without the data type extensions that will introduced later. Algorithm \ref{alg:ReliefF} details ReliefF as it has been efficiently implemented in ReBATE. Specifically all RBAs in ReBATE have been structured into distinct stages, i.e. (Stage 1) pre-process the data, (Stage 2) pre-compute the pairwise instance distance array, and (Stage 3) neighbor determination and calculate feature weights. 

The reasoning behind pre-computing the distance between all pairs of instances in Stage 2 is based on an assumption that was introduced in ReliefF and has largely persisted in most RBA implementations.  Specifically we assume that all training instances will be utilized in scoring.  This assumption is in contrast to the original Relief algorithm where the user could specify a subset of $m$ random instances that would be used to update feature weights \citep{kira1992practical}. However since it was found that the quality of weight estimates becomes more reliable as the parameter $m$ approaches the total number of instances $n$, proposed the simplifying assumption that $m=n$ \citep{kononenko1994estimating}. In other words, every instance gets to be the `target' for weight updates one time, i.e instances are selected without replacement. As such, all pairwise distances will be required to run each RBA in ReBATE.

\begin{algorithm}[h!]
\caption{Pseudo-code for ReliefF algorithm as implemented in ReBATE}
\label{alg:ReliefF}
\begin{algorithmic}[1]
\REQUIRE for each training instance a vector of feature values and the class value
\STATE \textit{n} $\gets$ number of training instances
\STATE \textit{a} $\gets$ number of attributes (i.e. features) 
\STATE \textbf{Parameter:} $\textit{k} \gets$ number of nearest hits `$H$' and misses `$M$' \\
\STATE
\STATE \emph{\textbf{\# STAGE 1}}
    \STATE pre-process dataset \COMMENT{$\approx a \cdot n$ time complexity} 
\STATE \emph{\textbf{\# STAGE 2}}
\STATE pre-compute distance array \COMMENT{$\approx 0.5 \cdot a \cdot n^{2}$ time complexity} 
\STATE \emph{\textbf{\# STAGE 3}}
\STATE initialize all feature weights $W[A]:=0.0$
\FOR{$i$:=1 \TO $n$}
    \STATE \emph{\textbf{\# IDENTIFY NEIGHBORS}}
    \FOR{j:=1 \TO $n$}
        \STATE identify $k$ nearest hits and $k$ nearest misses (using distance array)
    \ENDFOR
    \STATE \emph{\textbf{\# FEATURE WEIGHT UPDATE}}
    \FORALL{hits and misses}
        \FOR{A:= \TO $a$} 
            \STATE $W[A]:= W[A] - \text{\emph{diff}}(A,R_{i},H)/(n\cdot k)+\text{\emph{diff}}(A,R_{i},M)/(n\cdot k)$
        \ENDFOR
    \ENDFOR
\ENDFOR \\
\RETURN the vector $W$ of feature scores that estimate the quality of features
\end{algorithmic}
\end{algorithm}

All RBAs calculate a proxy statistic for each feature that can be used to estimate feature `relevance' to the target concept (i.e. predicting endpoint value). These feature statistics are referred to as feature weights ($W[A] =$ weight of feature `$A$'), or more casually as feature `scores' that can range from $-1$ (worst) to $+1$ (best). 

As depicted in Algorithm \ref{alg:ReliefF}, once ReliefF has pre-processed the data and pre-computed the distance between all instance pairs, Stage 3 cycles through $n$ randomly ordered training instances ($R_{i}$), selected without replacement. Each cycle, $R_{i}$ is the target instance and the feature score vector \emph{W} is updated based on feature value differences observed between the target and neighboring instances. ReliefF relies on a `number of neighbors' user parameter $k$ that specifies the use of $k$ nearest hits and $k$ nearest misses in the scoring update for each target instance. Next, it selects $k$ nearest neighbors with the same class called the \emph{nearest hits} ($H$) and the other with the opposite class, called the \emph{nearest misses} ($M$). The last step of the cycle updates the weight of a feature $A$ in $W$ if the feature value differs between the target instance $R_{i}$ and any of the nearest hits $H$ or nearest misses $M$. Features that have a different value between $R_{i}$ and an $M$ support the inference that they are informative of outcome, so the quality estimation \emph{W}[\emph{A}] is increased. In contrast, features with different between $R_{i}$ and $H$ suggest evidence to the contrary, so the quality estimation \emph{W}[\emph{A}] is decreased. The \emph{diff} function in Algorithm \ref{alg:ReliefF} calculates the difference in value of feature $A$ between two instances $I_{1}$ and $I_{2}$ (where $I_{1}=R_{i}$ and $I_{2}=$ either $H$, or $M$ when performing weight updates) \citep{robnik2001comprehensible}. For discrete (e.g. categorical or nominal) features, \emph{diff} is defined as:

\begin{equation}\label{eq:disc}
\text{\emph{diff}}(A,I_{1},I_{2}) =
\begin{cases}
0 & \text{if } value(A,I_{1})=value(A,I_{2})\\
1 & \text{if } otherwise\\
\end{cases}
\end{equation} 

\noindent and for continuous (e.g. ordinal or numerical) features, \emph{diff} is defined as:

\begin{equation}\label{eq:cont}
\text{\emph{diff}}(A,I_{1},I_{2}) = \frac{|value(A,I_{1})-value(A,I_{2})|}{max(A)-min(A)}
\end{equation} 

\noindent This function ensures that weight updates fall between 0 and 1 for both discrete and continuous features.  ReBATE adopts this strategy introduced by Relief \citep{kira1992practical} to extend all ReBATE algorithms for continuous features. Additionally, in updating \emph{W}[\emph{A}], (see line 19 of Algorithm \ref{alg:ReliefF}) dividing the output of \emph{diff} by $n$ and $k$ guarantees that all final weights will be normalized within the interval $[-1,1]$ adjusting for any level of class imbalance. This \emph{diff} function is further applied in pre-computing the distance array, calculating Manhattan distances between instance pairs. For efficiency, ReBATE pre-normalizes any continuous variable (i.e. features or endpoint) so that it falls within a 0 to 1 value range. 

Consider that while the above \emph{diff} function performs well when features are either uniformly discrete or continuous, it has been noted that given a dataset with a mix of discrete and continuous features, this \emph{diff} function can underestimate the quality of the continuous features \citep{kononenko2008non}. One proposed solution to this problem is a \emph{ramp function} that naively assigns a full \emph{diff} of 0 or 1 if continuous feature values are some user defined minimum or maximum value apart from one another \citep{hong1997use,robnik2003theoretical,kononenko2008non}. However this naive approach adds two additional user-defined parameters requiring problem-specific optimization.

\begin{figure}[t!]
	\centerline{\includegraphics[width=0.9\textwidth]{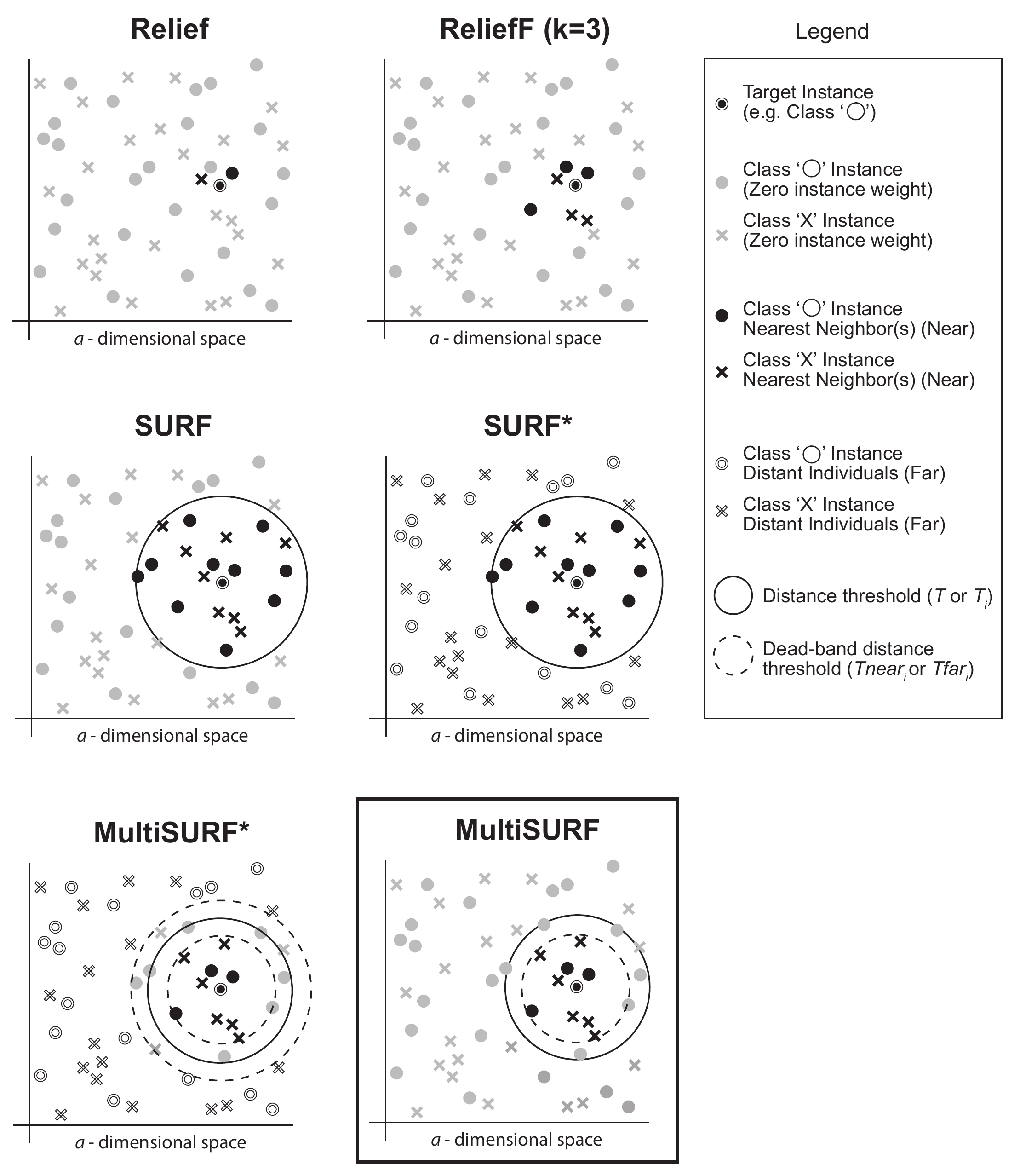}}
	\caption{Illustration of the neighbor selection differences between Relief, ReliefF, SURF, SURF*, MultiSURF*, and MultiSURF. Differences include the number of nearest neighbors or the method for selecting `near' or `far' instances for feature scoring. Note that for ReliefF, a $k$ of 3 is chosen but a $k$ of 10 is most common. These illustrations are conceptual and are not drawn to scale. }
	\label{fig:methods}
\end{figure}

Figure \ref{fig:methods} illustrates the major algorithmic differences between the original Relief algorithm \citep{kira1992practical}, ReliefF, and the four other core RBAs implemented in ReBATE. Specifically, this figure focuses on respective strategies for neighbor selection. Note that while a $k$ of 10 has been widely adopted as the default setting, a $k$ of 3 was chosen for this conceptual illustration.

\subsubsection{SURF} \label{SURF}
The SURF algorithm \citep{greene2009spatially} inherits the majority of the ReliefF algorithm. In contrast with ReliefF, SURF eliminates the user parameter $k$, instead adopting a distance threshold $T$ to determine which instances will be considered neighbors (see Figure \ref{fig:methods}). The radius defining the $T$ hyper-circle in  $a$-dimensional space around a given target instance is defined by the average distance between all instance pairs in the training data. This radius is therefore of a uniform size for each target instance.

\subsubsection{SURF*} \label{SURF*}
The SURF* algorithm \citep{greene2010informative} inherits the majority of the SURF algorithm. In contrast with SURF, SURF* introduced the concept of instances that were near vs. far from the target instance (see Figure \ref{fig:methods}). Applying the same $T$ from SURF, any instance within the threshold was considered near, and those outside were far. SURF* proceeds to weight `far' instance differences in an opposite manner than `near' instances. Specifically, feature value differences in hits differently receive a ($+1$) while feature value differences in misses differently receive a ($-1$). Notably, SURF* is the only RBA we implement that uses all instances in the training dataset to update feature weights each cycle. Note that the '*' in naming signifies the use of `far' scoring in both SURF* and MultiSURF*.

\subsubsection{MultiSURF*} \label{MultiSURF*}
The MultiSURF* algorithm \citep{granizo2013multiple} inherits the majority of the SURF* algorithm. In contrast with SURF*, MultiSURF* defines a threshold $T_{i}$ as the mean pairwise distance between the target instance and all others, as opposed to the mean of all instance pairs in the data. This adapts the definition of near/far to a given part of the feature space. MultiSURF* also introduces a dead-band zone extending on either side of $T_{i}$, i.e. $Tnear_{i}$ or $Tfar_{i}$ (see Figure \ref{fig:methods}) to exclude instances near $T_{i}$ from contributing to scoring, i.e. those that are ambiguously near or far.  Accordingly, the width of the dead-band zone is the standard deviation $\sigma$ of pairwise distances between the target instance and all others. Thus, for each target instance, the boundary circles, as illustrated for MultiSURF* in Figure \ref{fig:methods}, may have different radii. Lastly, the `far' scoring logic was inverted in to reduce computations. Specifically in SURF*, \emph{differences} in feature values in hits yielded a reduction in feature score, and an increase in misses. Since differences are expected to be more frequent in far individuals, MultiSURF* updates far instance feature weights with the \emph{same} feature values, i.e. hits receive a ($+1$), and misses receive a ($-1$).

\subsubsection{MultiSURF} \label{MultiSURF}
The new RBA variant proposed in this study is called MultiSURF. As the minor name change suggests, MultiSURF is closely related to MultiSURF* \citep{granizo2013multiple}. MultiSURF preserves all aspects of MultiSURF* but eliminates the `far' scoring introduced in SURF* (see Figure \ref{fig:methods}).  The dead-band boundary $Tnear_{i}$ in MultiSURF is equal to $T_{i}-\sigma_{i}/2$.  Pseudo-code for MultiSURF is given by Algorithm \ref{alg:MultiSURF}, again excluding the data type expansions for simplicity. As before, the \emph{diff} function of MultiSURF is given by Equations \ref{eq:disc} and \ref{eq:cont}.  Also note that because neighbors are defined by a threshold in MultiSURF, there can be a variable, or imbalanced number of hits and misses for each target instance. Lines 15 to 25 of Algorithm \ref{alg:MultiSURF} identify nearest hits and misses and track counts of each ($h$ and $m$, respectively). Line 29 normalizes weight updates based on $n$, $h$, and $m$. This accounts for imbalanced hit and miss counts for a given target instance. By using either an equal number of hits or misses (e.g. as in ReliefF), or using this type of $h$ and $m$ normalization, RBAs inherently adjusts for class imbalance in binary or multi-class problems \citep{robnik2003experiments}. As it is with all other core RBAs, the asymptotic time complexity of MultiSURF is $\mathcal{O}(n^2\cdot a)$. The complete time complexity of MultiSURF is $c_{0}a  +  c_{1} 0.5 n^2 a  + c_{2} n \log n + c_{3} 0.31n^2a + c_{y} n^2$ which is slightly faster than for MultiSURF*. Time complexity comparisons with of other RBAs are detailed in \citep{urbanowicz2017relief}.

\begin{algorithm}[h!]
\caption{Pseudo-code for the proposed MultiSURF algorithm in ReBATE}
\label{alg:MultiSURF}
\begin{algorithmic}[1]
\REQUIRE for each training instance a vector of feature values and the class value
\STATE \textit{n} $\gets$ number of training instances
\STATE \textit{a} $\gets$ number of attributes (i.e. features) \\
\STATE
\STATE \emph{\textbf{\# STAGE 1}}
    \STATE pre-process dataset \COMMENT{$\approx a \cdot n$ time complexity} 
\STATE \emph{\textbf{\# STAGE 2}}
\STATE pre-compute distance array \COMMENT{$\approx 0.5 \cdot a \cdot n^{2}$ time complexity} 
\FOR{$i$:=1 \TO $n$}
    \STATE set $T_{i}$ to mean distances between instance $i$ and all others
    \STATE set $\sigma_{i}$ to standard deviation of those distances
\ENDFOR
\STATE \emph{\textbf{\# STAGE 3}}
\STATE initialize all feature weights $W[A]:=0.0$
\FOR{$i$:=1 \TO $n$}
    \STATE \emph{\textbf{\# IDENTIFY NEIGHBORS}}
    \STATE initialize hit and miss counters $h:=0.0$ and $m:=0.0$
    \FOR{j:=1 \TO $n$}
        \IF{distance between $i$ and $j$ is $<$ $T_{i}$ - $\sigma_{i} /2$ (using distance array)} 
            \IF{$j$ is a hit} \STATE{$h+=1$} \COMMENT{and identify instance as hit} \ELSIF{$j$ is a miss} \STATE{$m+=1$} \COMMENT{and identify instance as miss} \ENDIF
        \ENDIF
    \ENDFOR
    \STATE \emph{\textbf{\# FEATURE WEIGHT UPDATE}}
    \FORALL{hits and misses}
        \FOR{A:= \TO $a$} 
            \STATE $W[A]:= W[A] - \text{\emph{diff}}(A,R_{i},H)/(n \cdot h)+\text{\emph{diff}}(A,R_{i},M)/(n\cdot m)$
        \ENDFOR
    \ENDFOR
\ENDFOR \\
\RETURN the vector $W$ of feature scores that estimate the quality of features
\end{algorithmic}
\end{algorithm}

\subsubsection{Multi-class Endpoint}
In the remaining subsections, we describe how RBAs in ReBATE were extended to handle respective data-type issues. To address multi-class endpoint problems, ReBATE methods adopts a strategy similar to one proposed by \citet{robnik2003theoretical} for ReliefF that finds $k$ nearest misses from every `other' class, and averages the weight update based on the prior probability of each class. This implementation would requires the calculation of prior class probabilities P(C) from the training data and, for a given target instance, identifying misses of every \emph{other} class. Equation \ref{eq:multi-class:ReliefF} defines this multi-class update equation.

While this approach works fine for ReliefF, where an equal `$k$' number of neighbors are selected for each miss class, in all of our other RBAs, the radius strategy leaves open the possibility that one or more miss classes will not fall within the radius, that thus not be included in the scoring update. As this could lead to a scoring bias we have implemented a simpler strategy that is appropriate for all ReBATE methods. Instead of normalizing miss class contributions by a factor of P(C), they are normalized by the proportion of each miss class within the set of selected neighbors. This proportion is simply calculated as the number neighbors from the current `miss' class ($m_{C}$ divided by the total number of neighbors that were `misses' ($m$). Equation \ref{eq:multi-class:ReliefFNEW} defines our new multi-class weight update equation for ReliefF and Equation \ref{eq:multi-class:MultiSURF} does the same for SURF, MultiSURF and for `near' neighbors in SURF* and MultiSURF*. For `far' instances in SURF* and MultiSURF*, the update would look like Equation \ref{eq:multi-class:MultiSURF}, however the hit term would be added and the summed miss term would be subtracted. 

\begin{equation}\label{eq:multi-class:ReliefF}
\begin{split}
W[A] &:= W[A] - \text{\emph{diff}}(A,R_{i},H)/(n*k)+ \\
& \sum_{C\neq class(target)}\Big[\frac{P(C)}{1-P(class(target))} \text{\emph{diff}}(A,R_{i},M(C))\Big]/(n*k) 
\end{split}
\end{equation} 

\begin{equation}\label{eq:multi-class:ReliefFNEW}
\begin{split}
W[A] &:= W[A] - \text{\emph{diff}}(A,R_{i},H)/(n*k)+ \\
& \sum_{C\neq class(target)}\Big[\frac{m_{C}}{m} \text{\emph{diff}}(A,R_{i},M(C))\Big]/(n*k) 
\end{split}
\end{equation} 

\begin{equation}\label{eq:multi-class:MultiSURF}
\begin{split}
W[A] &:= W[A] - \text{\emph{diff}}(A,R_{i},H)/(n*h)+ \\
& \sum_{C\neq class(target)}\Big[\frac{m_{C}}{m} \text{\emph{diff}}(A,R_{i},M(C))\Big]/(n*m) 
\end{split}
\end{equation} 

\subsubsection{Regression}
To address regression, we propose an alternative, simpler approach than the current standard introduced in the Regressional ReliefF (RReliefF) algorithm \citep{kononenko1996relieff,robnik1997adaptation}. The fundamental challenge of adapting Relief algorithms to continuous endpoints, is that we lose a clear definition for hit or miss, i.e. having the same or different class. RReliefF proposed a kind of ``probability'' that two instances belong to two ``different'' classes. This ``probability'' is modeled with the distance between feature and endpoint values of two learning instances as detailed by \citet{robnik1997adaptation}. This includes an exponential weighting of instance contributions to W[A] based on distance between instances. Since current ReBATE methods do not apply distance based instance weights, and RReliefF requires an additional step computing prior and conditional probabilities, we propose a simpler regression scheme for our ReBATE methods. 

Specifically, ReBATE calculates the standard deviation of the continuous endpoint ($\sigma_{E}$) and applies this as a simple threshold for determining whether two instances will be considered a ``hit'' or a ``miss''. This serves to contextually discretize the continuous endpoint into `same class' or `different class' from the perspective of the target instance. This proposed adaptation of RBAs to regression problems only requires pre-computing $\sigma_{E}$ during Stage 1 of the algorithm, and changing the definition of a hit from $C_{i} = C_{j}$ to $|C_{i} - C_{j}| < \sigma_{E}$, and the definition of a miss from $C_{i} \neq C_{j}$ to $|C_{i} - C_{j}| \geq \sigma_{E}$.

\subsubsection{Missing Data}
Missing feature values must be dealt with by RBAs at two points in the algorithm: (1) Calculation of distances between instance pairs and (2) updating the feature weights. Previously, a missing data strategy proposed in ReliefF (or more precisely in ReliefD) had been identified as `best' with minimal empirical investigation \citep{kononenko1994estimating}. It was also designed explicitly for problems with discrete endpoints. Specifically, depending on whether one or both instances have a missing value for the given feature, the \emph{diff} function returns the probability that the feature states are different given the class of each instance. This approach is implicitly a form of interpolation, making an `educated' guess at what the missing value might be. Under the right circumstances, this can indeed improve performance, but if the guess is wrong, it could just as easily harm performance. Further, this approach is more computationally and conceptually challenging to extend to continuous endpoint data.

In ReBATE we propose what we call an `agnostic' approach to missing data that is most similar to one considered in ReliefC \citep{kononenko1994estimating}. The idea behind an agnostic approach is that unknown, missing values should be ignored (i.e. treated neutrally) using normalization to bypass their inclusion rather than attempt to make a guess about their respective values.  In contrast, the ReliefC method is only partially agnostic, in that uses the ReliefB method \citep{kononenko1994estimating} to naively contribute a \emph{diff} of $1- \frac{1}{\#Unique\_Feature\_Values}$ when a missing value is encountered in calculating the distance between an instance pair \citep{kononenko1994estimating}. For example, this contribution would be 0.5 if the feature had two possible states, or 0.25 if it had four. However, when ReliefC updates feature weights, features with missing values contribute nothing and the distance score is normalized to reflect that it was calculated using ($a-\#Missing\_Features$), where $\#Missing\_Features$ is the number of features where a missing value was observed for at least one of the two instances. Alternatively, ReBATE methods apply this agnostic treatment of missing data to both the calculation of instance pair distances as well as for feature weight updates. This approach easily integrates with all RBAs, and all other data-type extensions. To the best of our knowledge this study is the first to implement and test a fully agnostic missing data approach in RBAs.

\subsection{Evaluation}
In the present study, we compare 13 feature selection approaches over an archive of 2280 simulated datasets representing a variety of problem and data types. In addition to the 5 ReBATE algorithms already reviewed or described above (ReliefF, SURF, SURF*, MultiSURF*, and MultiSURF) we examine 3 alternate run settings for ReliefF as well as 5 established non-RBA feature selection methods.  

\subsubsection{ReliefF Runs}
The different ReliefF runs will be labeled in the results as: ReliefF 10 NN (i.e. original ReliefF), ReliefF 100 NN, ReliefF 10\% NN, and ReliefF 50\% NN. The first two are ReliefF with a $k$ of 10 or 100, respectively.  The second two consider setting $k$ in a dataset dependent manner, setting $k$ based on a user defined percent of instances in the data. For example, if $n$ were 1000 instances, ReliefF 10\% NN would utilize 100 total instances, thus $k=50$, i.e. 50 hits and 50 misses.  

We explore these different settings of $k$ in ReliefF to explore how the number of nearest neighbors impacts performance, as well as whether setting $k$ based on a percentage of instances offers a potential alternative to threshold-based neighbor selection as used by SURF, SURF*, MultiSURF*, and MultiSURF. 

\subsubsection{Other Feature Selection Algorithms}
We compare the ReBATE methods to a cross section of feature selection methods available in scikit-learn \citep{scikit-learn}. Specifically we compare to three established filter methods including the chi squared test \citep{vafaie1994feature,zheng2004feature,jin2006machine,moore2007tuning,witten2016data}, ANOVA F-value \citep{guyon2003introduction,forman2003extensive,jafari2006assessment,jovic2015review}, and mutual information (i.e. information gain) \citep{hunt1966experiments,vergara2014review,hoque2014mifs}. Like most filter methods, these methods are myopic. Myopic methods evaluate features independently without considering the context of other features \citep{kononenko2008non}. Thus, they are not be expected to detect feature dependencies due to assumptions of variable independence.  Notably, the chi squared test is limited to problems with a discrete endpoint, and the ANOVA F-value was selected for its applicability to multi-class endpoints. 

Further, we compare two wrapper methods, each based on a random forest of decision trees, i.e. (1) ExtraTrees, and (2) RFE ExtraTrees \citep{geurts2006extremely}. Random forests have been recognized as a powerful ensemble machine learning approach \citep{liaw2002classification}. For both algorithms, we utilized 500 estimators (i.e. number of trees in the forest), and left all other parameters to scikit-learn defaults. Feature importance scores output by either algorithm are used to rank features by potential relevance. RFE ExtraTrees is a random forest combined with a \emph{recursive feature elimination} algorithm. We have run RFE ExtraTrees removing one feature each step (the most conservative setting). RFE ExtraTrees is an iterative approach, recalculating feature importance of the remaining features each iteration. We expect this algorithm to be, by far, the most computationally expensive of those evaluated. RFE ExtraTrees would be more fairly compared to iterative versions of RBAs including TuRF, but we include it here to emphasize the comparative power of core RBA methods.

\subsubsection{Simulation Study}
In previous comparisons of ReliefF, SURF, SURF*, and MultiSURF* these methods were evaluated on datasets with purely epistatic 2-way interactions (i.e. no main effects) with varying numbers of training instances (e.g. 200 to 3200) as well as different heritabilities (e.g. 0.01 to 0.4) \citep{greene2009spatially,greene2010informative,granizo2013multiple}. Heritability is a genetics term that indicates how much endpoint variation is due to the genetic features.  In the present context, heritability can be viewed as the signal magnitude, where a heritability of 1 is a `clean' dataset (i.e. with the correct model, endpoint values will always be correctly predicted based on feature values), and a heritability of 0 would be a completely noisy dataset with no meaningful endpoint associations. All features were simulated as single nucleotide polymorphisms (SNP) that could have have a discrete value of (0, 1, or 2) representing possible genotypes.  In each dataset, two features were predictive (i.e. relevant) of a binary class while the remaining 998 features were randomly generated, based on genetic guidelines of expected genotype frequencies, yielding a total of 1000 features.  Similarly, VLSRelief explored SNP simulations and 2-way epistasis varying heritability similar to the other studies, but fixing datasets to 1600 instances and simulating datasets with either 5000 or 100,000 total features \citep{eppstein2008very}. It should be noted that most of these studies sought to compare core RBAs to respective iterative TuRF expansions, which is why larger feature spaces were simulated.

Simulation studies such as these facilitate proper evaluation and comparison of methodologies because a simulation study can be designed by systematically varying key experimental conditions, and the ground truth of the dataset is known i.e. we know which features are relevant vs. irrelevant, we know the pattern of association between relevant features and endpoint, and we know how much signal is in the dataset (so we know what testing accuracy should be achievable in downstream modeling). This allows us to perform power analyses over simulated dataset replicates to directly evaluate the success rate of our methodologies. 

For these reasons, we adopt the position that the best way to compare and evaluate machine learning methodologies is over a diverse panel of simulated datasets designed to ask generalizable questions about what a method can and cannot handle (e.g. 2-way epistatic interactions, missing data, high noise, a moderate sample size (800), etc.). Ultimately, it may not be very important that the same benchmark datasets are used across studies, but rather that simulation studies are designed to ask fundamental questions about generalizable methodological functionality.  

Therefore in the present study we have designed and applied a simulation study that is inspired by, but goes well beyond, the other bioinformatic evaluations previously described.  Broadly speaking, our simulation study is founded around a core set of pure 2-way interaction SNP datasets similar to those previously benchmarked, but we expand beyond these to include groups of SNP datasets with (1) a variety of simple main effects, (2) 3-way interactions, (3) genetic heterogeneity, (4) continuous-valued features, (5) a mix of discrete and continuous features, (6) multi-class endpoints, (7) continuous endpoints, (8) missing data, and (9) imbalanced data. Additionally, we include some clean toy benchmark datasets including the XOR problem (2-way to 5-way interactions) to explore higher order interactions, and the multiplexer problem (6-bit to 135 bit variations) to explore epistasis and heterogeneous associations simultaneously.  

Table \ref{tab:data} breaks down the characteristics of each unique dataset group. Outside of the core datasets, most of the other dataset groups retain constraint settings known to be solvable among the core datasets. These include the inclusion of a 2-way interaction, 20 features, a heritability of 0.4, and 1600 training instances. This is done because it would be very computationally expensive to evaluate a full factorial set of dataset variations over all dataset constraints. 

\begin{table*}[h!]
 \centering
\caption{Simulation study datasets. 30 replicates of each configuration were generated. Model architecture difficulties are designated by `E' (easy), and `H' (hard). Simulation method generation is designated as either `G' (GAMETES), `C' (custom script), or `G+C' (GAMETES modified by custom script).}
\label{tab:data}       
{\scriptsize 
\begin{tabular}{|l|c|c|c|c|c|c|c|c|}
\hline\
 & \multirow{9}{*}{\rotatebox{90}{\textbf{Configurations}}} & \multirow{9}{*}{\rotatebox{90}{\textbf{Config. Variations}}} & \multirow{9}{*}{\rotatebox{90}{\textbf{Predictive Features}}} & \multirow{9}{*}{\rotatebox{90}{\textbf{Total Features}}} & \multirow{9}{*}{\rotatebox{90}{\textbf{Model Difficulty}}}& \multirow{9}{*}{\rotatebox{90}{\textbf{Heritability}}} & \multirow{9}{*}{\rotatebox{90}{\textbf{Instances}}} & \multirow{9}{*}{\rotatebox{90}{\textbf{Simulation Method}}} \\
& & & & & & & &  \\ 
& & & & & & & &  \\ 
& & & & & & & &  \\ 
& & & & & & & &  \\ 
\textbf{Simulated Data Group}& & & & & & & &  \\ 
\textbf{Description or} & & & & & & & &  \\ 
\textbf{Pattern of}& & & & & & & &  \\ 
\textbf{Association} &  & & & & & & &  \\ \hline \hline

\rowcolor{gray!30}2-way Pure Epistais  & 32 & - & 2 & 20 & E, & 0.05,& 200, & G  \\
\rowcolor{gray!30}(Core Datasets) & & & & & H & 0.1,& 400,& \\
\rowcolor{gray!30} & & & & && 0.2,& 800,& \\
\rowcolor{gray!30}\emph{Others marked by `*'}& & & & && 0.4 & 1600 & \\ \hline
 
1-Feature Main Effect & 8 & - & 1 & 20 & E, & 0.05,& 1600 & G   \\
& &  & & & H & 0.1,& & \\
& & & & && 0.2,& & \\
& & & & && 0.4 & & \\ \hline

\rowcolor{gray!30}2-Feature Additive Effect & 2 & 50:50, & 2 & 20 & E & 0.4 & 1600 & G  \\
\rowcolor{gray!30}& & 75:25 & & & & & & \\ \hline

4-Feature Additive Effect & 1 & - & 1 & 20 & E& 0.4 & 1600 & G \\ \hline

\rowcolor{gray!30}4-Feat. Additive  & 2 & 50:50, & 2 & 20 & E& 0.4 & 1600 & G  \\
\rowcolor{gray!30}2-way Epistasis & & 75:25 & & & & & & \\ \hline

4-Feat. Heterogeneous  & 2 & 50:50, & 2 & 20 & E& 0.4 & 1600 & G \\
2-way Epistasis & & 75:25 & & & & & & \\ \hline

\rowcolor{gray!30}3-way Pure Epistasis & 1 & - & 3 & 20 & E& 0.2 & 1600 & G  \\ \hline

Number of Features* & 4 & - & 2 & 100, & E& 0.4 & 1600 & G  \\
 & & & &  1000,& & & & \\
& & & &  10000, & & &  & \\
& & & & 100000 & & & & \\ \hline

\rowcolor{gray!30}Continuous Features* & 1 & - & 2 & 20 & E& 0.4 & 1600 & G+C \\ \hline

Mix of Discrete and  & 1 & - & 2 & 20 & E& 0.4 & 1600 & G+C \\ 
Continuous Features* & & & & & & & & \\ \hline

\rowcolor{gray!30}Continuous Endpoint* & 3 & 0.2, 0.5, 0.8 & 2 & 20 & E & 0.4 & 1600 & G  \\  \hline

Continuous Endpoint* & 1 & - & 2 & 20 & E & 0.4 & 1600 & G+C  \\ 
(1-Threshold Model) & & & & & & & & \\ \hline

\rowcolor{gray!30}Missing Data* & 4 & 0.001, 0.01, & 2 & 20 & E & 0.4 & 1600 & G+C  \\ 
\rowcolor{gray!30} & & 0.1, 0.5 & & & & & & \\ \hline

Imbalanced Data* & 2 & 0.6, 0.9 & 2 & 20 & E & 0.4 & 1600 & G  \\  \hline

\rowcolor{gray!30}Multi-class Endpoint & 2 & 3-class, & 2 & 20 &N/A& 1 & 1600 & C  \\
\rowcolor{gray!30}(Impure 2-way Epistasis) & & 9-class & & &  && & \\ \hline

XOR Model & 4 & 2-way, & 2 & 20 & N/A & 1 & 1600 & C   \\
(Pure Epistasis) &  & 3-way,& 3 & & & & & \\
& & 4-way, & 4 & & & & & \\
& & 5-way & 5 & & & & & \\ \hline \hline

\rowcolor{gray!30}Multiplexer (MUX) & 6 & 6-bit $\rightarrow$ & 2 & 6 & 3-way & 1 & 500 & C   \\
\rowcolor{gray!30}(Pure Epistasis and& & 11-bit $\rightarrow$ & 3 & 11& 4-way& & 1000 & \\
\rowcolor{gray!30}Heterogeneous  & & 20-bit $\rightarrow$ & 4 & 20& 5-way& & 2000 & \\
\rowcolor{gray!30}Associations) & & 37-bit $\rightarrow$ & 5 & 37& 6-way& & 5000 & \\ 
\rowcolor{gray!30} & & 70-bit $\rightarrow$ & 6 & 70& 7-way& & 10000& \\
\rowcolor{gray!30}& & 135-bit $\rightarrow$ & 7 & 135& 8-way& & 20000 & \\ \hline

\end{tabular}
} 
\end{table*}

Any dataset group in Table \ref{tab:data} that simulates a 2-way pure epistatic interaction is marked by a `*'. The left-most column describes the generalized pattern of association or data type that might reflect a strength or weakness we wish to assess.  The `Configurations' column indicates the number of unique dataset configurations that were included in the respective group.  For example, the first group of `core datasets' includes 32 configurations, i.e. 2 model difficulties * 4 heritabilities * 4 instance counts. The `Model Difficulty' refers the model architecture of the underlying simulated genetic model \citep{urbanowicz2012predicting}. To capture this dimension of dataset complexity, we select models generated at the extremes of model difficulty labeled here as `easy' (E) and `hard' (H). The `Config. Variation' column is a catchall for configuration variations in a given group. The `Simulation Method' refers to the strategy used to generate the datasets.  Datasets simulated with the GAMETES complex genetic model and dataset generation software \citep{urbanowicz2012gametes,urbanowicz2012predicting} are labeled with `G'. Those generated by a custom script are labeled with `C'. Those generated by GAMETES but later modified (e.g. discrete values transformed into a range of continuous values) are labeled with `G+C'. 

To clarify specific dataset groups, \emph{2-feature additive effect} includes two features with main effects that are additively combined to determine endpoint. The ratio 50:50 indicates that both features equally influenced endpoint, while 75:25, indicates that one had a 3 times the influence (and thus the relevance) of the other. Regarding \emph{4-feature additive effect}, all four features contribute equally to endpoint. The group, \emph{4-feature additive 2-way epistasis}, additively combines two distinct 2-way pure epistatic interactions, where each pair has the respective ratio of influence. The group, \emph{4-feature heterogeneous 2-way epistasis}, generates a heterogeneous pattern of association between two independent 2-way pure epistatic interactions. This would be an example of simulated genetic heterogeneity since these are SNP datasets. The group, \emph{continuous endpoints} is an example of a regression problem, often known in genetics as a quantitative trait endpoint. We apply the GAMETES software \citep{urbanowicz2012gametes} to generate quantitative trait values around each pair-wise genotype combination with a standard deviation of either 0.2, 0.5, or 0.8. For these datasets, the effective heritability is degraded as the standard deviation setting increases. The group, \emph{continuous endpoint} with a 1-threshold model, refers to an alternative approach to generating continuous endpoint datasets. This approach takes a SNP dataset generated by GAMETES software and converts any instance with a class of 0 to a random value between 0 and 50, and any with a class of 1 to a random value between 50 and 100.  This creates a continuous endpoint scenario, where a meaningful quantitative threshold exists in the data (50 in this case). Notably, we use a similar approach to generate our continuous-valued features, and mixed discrete/continuous feature datasets.  However for these, SNP values of 0, 1, or 2 are converted to random value between 0 and 50, 50 and 100, and 100 and 150, respectively. For \emph{missing data}, different frequencies of 'N/A's' are added to respective core datasets. For \emph{imbalanced data} the given class imbalance ratios are simulated. For \emph{multi-class endpoint} we simulated SNP datasets with a model similar to the XOR model, however each 2-way genotype combination is assigned either one of 3 classes or one of 9-classes.  In both situations datasets are generated with impure epistatic interactions (meaning that individual features each also have some main effect). 

Lastly, we have included six multiplexer problem datasets. Multiplexer problems are detailed in \citep{urbanowicz2017introduction}. In summary, they are clean problems with binary feature values, and a binary endpoint that concurrently model a patterns of epistastis and heterogeneous associations. We set these datasets apart in the table, because each dataset has a unique set of characteristics. For instance, the 20-bit multiplexer has 20 total features, involves heterogeneous groups of 5-way pure epistastic interactions, and includes a sample size of 2000.  Notably in all multiplexer problems, all the features are technically predictive (in at least some subset of the training data). However, for any multiplexer problem, specific features known as `address bits' are predictive in every instance. We specify the number of address bits under the `Predictive Features' column and the order of epistatic interaction in the `Model Difficulty' column. Previously it was noted in \citep{urbanowicz2015exstracs} that properly prioritizing address bits over other features in feature weighting with MultiSURF* was key to solving the 135-bit multiplexer problem directly. Therefore, with respect to the multiplexer problems, we evaluate the ability of feature selection methods to rank address bits as predictive features above all others (regardless of the fact that all features are technically predictive). 

In total we consider 76 unique dataset configurations, generating 30 randomly seeded replicate datsets for each configuration ($76*30 = 2280$ total datasets). We have made these datasets available for download\footnote{https://github.com/EpistasisLab/rebate-benchmark}. There are certainly many other dataset variations that could be included in the future, but this proposed set represents the most diverse simulation study of RBAs to date, offering a broad snapshot of the basic strengths and weaknesses of the feature selection methods evaluated in this study. 

\subsubsection{Analysis}
The strategy for evaluating a feature selection approach can depend on whether it outputs a ranked feature list (i.e. individual evaluation filter approaches), or a specific feature subset (all methods can do this).  Assuming a ranked feature list and a dataset where the ground truth is known ahead of time, it is most common to examine where all relevant features rank in the ordered feature score list (i.e. are they at the very top or among some top percentile?) \citep{kira1992practical,florez2002reviewing,moore2007tuning,eppstein2008very,greene2009spatially,greene2010informative,stokes2012application,granizo2013multiple}. Ideally, relevant features will all have higher scores than irrelevant features as a best case scenario but it is most important that relevant features at least make it above the selected feature subset cutoff. Other metrics including \emph{separability} and \emph{usability} have also been proposed \citep{robnik2003theoretical}. 

Alternatively, if the feature selection approach outputs a feature subset we can evaluate success by (1) examining the number of relevant and irrelevant features that comprise a selected feature subset (assuming ground truth is known) \citep{belanche2011review,bolon2013review}, or (2) determining the testing accuracy of some induction algorithm model trained on that feature subset (if ground truth is not known) \citep{sun2006iterative,bolon2013review}. The downside to the second approach is that it is difficult to separate the performance of the feature selection approach from the modeling of the induction algorithm. If we are dealing with an individual evaluation approach that has employed a selection cutoff to define a feature subset, then the downside to the first approach is that we are evaluating the feature weighting as well as the cutoff criteria (which can also be difficult to separate).  

For individual evaluation filter approaches like RBAs, the best way to evaluate performance on simulation study data where the ground truth is known, is to identify where the relevant feature rank in the ordered feature score list over a number of dataset replicates. This offers the clearest analysis for interpretation and is consistent with previous evaluations of the selected ReBATE methods. Specifically we apply a power analysis, examining where the lowest scoring of the relevant features ranks in the ordered feature list. Power, i.e. success rate, is then calculated as the proportion of successes out of the 30 replicate datasets. In this study, we calculate and report the power of each algorithm to identify all predictive features within each percentile of the ranked feature list.  

\begin{figure}[t]
	\centerline{\includegraphics[width=0.75\textwidth]{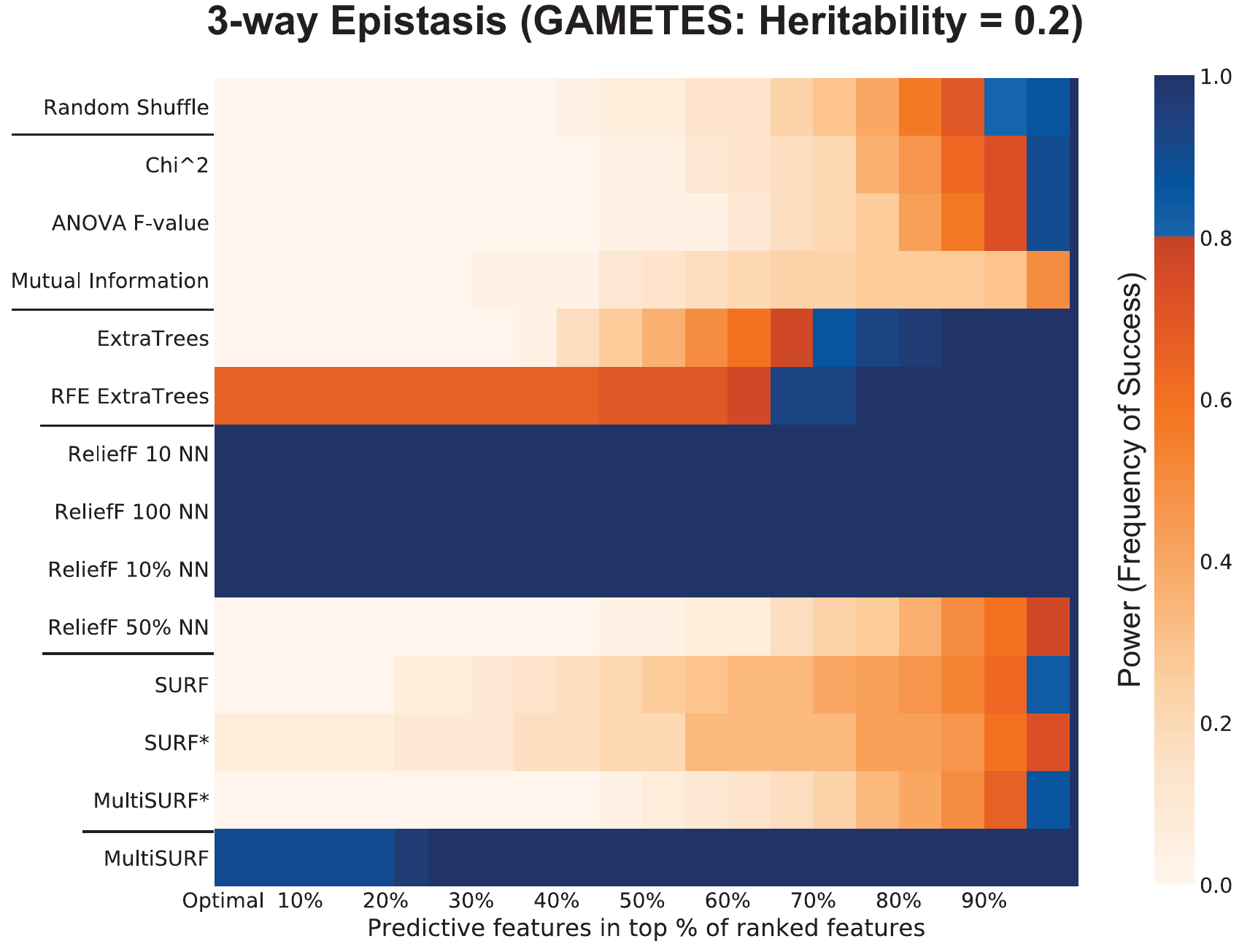}}
	\caption{This heatmap illustrates the power of different feature selection algorithms to rank all predictive features in the the top scoring `x' percent of features in the dataset. Results for the noisy 3-way epistatic interaction.}
	\label{fig:Results:3-way}
\end{figure}

As mentioned in this study, we compare 13 feature selection approaches across 76 unique dataset configurations. We have generated heatmaps displaying feature selection power at each percentile of a respective feature list. As an example, see Figure \ref{fig:Results:3-way}. On the y-axis we have our 13 algorithms, along with the results of a negative control, labeled as `Random Shuffle'. This negative control represents shuffling the feature list randomly 30 different times and calculating power. In other words, this is the power expected by randomly ranking features. The 13 feature selection algorithms are ordered and delineated into groups. From the top down, the first group includes the myopic filter-based feature selection methods (i.e. chi square, ANOVA F-value, and mutual information). The second group includes the random forest wrapper methods (i.e. ExtraTrees and RFE ExtraTrees). The third group includes the ReliefF algorithm with different settings of the $k$ parameter. The fourth group includes the set of recent RBAs that have eliminated the $k$ parameter, and the last algorithm is MultiSURF, proposed in this study. We will use lines delineating these algorithm groups in some figures where the text of the algorithm names would be too small to read, but the order of these algorithms is preserved for all remaining power analysis figures. 

The x-axis of Figure \ref{fig:Results:3-way} gives the percentile of the ranked feature scores (e.g. $30\%$ represents the top $30\%$ of all feature scores). The zero percentile is marked as `optimal'. At optimal, all relevant features are scored above all irrelevant features. To the right of the figure is the key depicting increasing power with increasing color darkness. To facilitate interpretation, any measure of power at or above 0.8 (i.e. $80\%$ power) is given as a shade of blue, rather than a shade of orange. While somewhat of an arbitrary selection, $80\%$ power or above is often used as a significance cutoff for success rate. It is useful here to more quickly identify the lowest feature percentile within which all relevant features are ranked with significant rate of success (i.e. the threshold between orange and blue). For example, if we look at the power of RFE ExtraTrees in Figure \ref{fig:Results:3-way}, we can see that this algorithm successfully ranks all relevant features somewhere in the top $67\%$ of features for at least $80\%$ of the replicate datasets. An algorithm that performs perfectly will have dark blue band ($100\%$ power) over the entire percentile range (e.g. see ReliefF 10 NN in Figure \ref{fig:Results:3-way}). This indicates that the algorithm succeeded in scoring all relevant features above each irrelevant feature in every replicate dataset. Keep in mind that when it comes to selecting a feature subset from a ranked feature list, the minimum basis for success is whether all predictive features are included within that set, not whether all relevant features are ranked above every irrelevant feature. For example if we had decided to keep the top $25\%$ of features in subset selection, then we would want to pick an algorithm that had reliable power at the 25th percentile. Lastly, notice that by the 100th percentile, all algorithms will report $100\%$ power. This is to be expected since we know that all relevant features will be found \emph{somewhere} in the entire feature set. A Jupyter notebook including our analysis code and power analysis figure generation has been made available\footnote{https://github.com/EpistasisLab/rebate-benchmark}.

\section{Results} \label{results}
The results of this study are organized by major data configuration themes. 

\subsection{2-way Epistasis}
\begin{figure}[t!]
	\centerline{\includegraphics[width=0.9\textwidth]{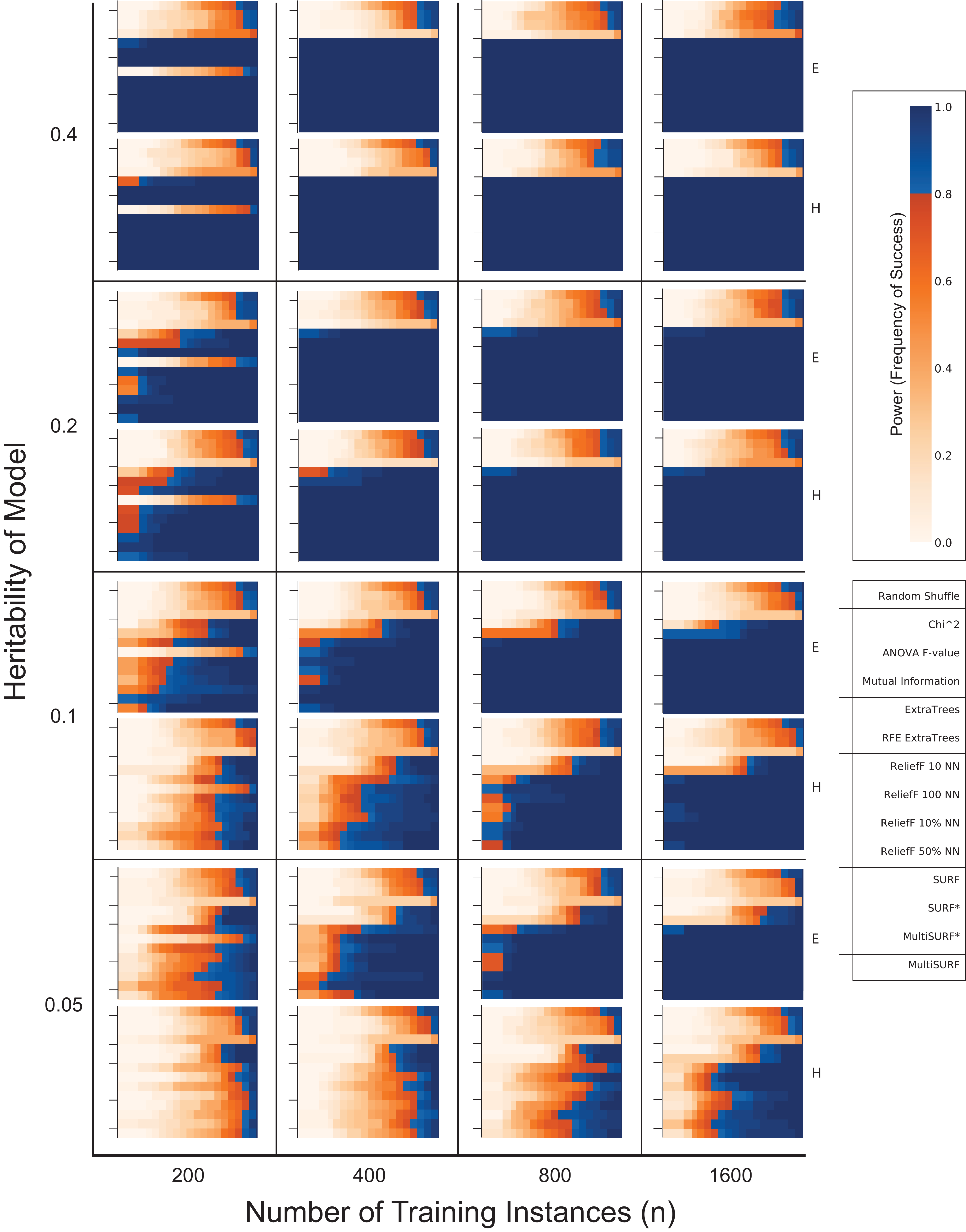}}
	\caption{Results for all core 2-way epistatic interaction datasets. Keys relevant to all plots are given on the far right. Tick marks delineating algorithm groups are provided for each sub-plot.}
	\label{fig:Results:2-way}
\end{figure}

Figure \ref{fig:Results:2-way} presents the results for the core set of 2-way pure epistastic interaction datasets. This figure assembles power plots over a range of heritabilities (left y-axis), number of instances (x-axis), and the two model architecture difficulties, E and H (right y-axis). Each of the 32 subplots represents a power analysis for one data configuration. They are arranged roughly so that the `easiest' configurations are towards the upper right corner (e.g. heritability $= 0.4$, $n=1600$, and architecture = E), and the most `difficult' configurations are towards the lower left corner (e.g. heritability $= 0.05$, $n=200$, and architecture = H). 

The most obvious observation from this analysis is that the three myopic filter algorithms (i.e. chi square, ANOVA F-test, and mutual information) consistently fail to successfully rank relevant features with 2-way interactions, i.e. their performance is on par with the random shuffle negative control. Also notice that overall power of all other algorithms deteriorates as we go from the upper right plot to the lower left plot. This is consistent with the the expectation that power will degrade with decreasing heritability, training set size, and a `harder' model architecture. Note how our simulation study design includes data configurations where our RBAs of interest completely succeed (upper right) as well as completely fail (lower left). These trends are consistent with previous evaluations of ReliefF (10 NN), SURF, SURF*, and MultiSURF* \citep{greene2009spatially,greene2010informative,granizo2013multiple}.

Focusing on the performance of ReliefF with different settings of $k$ one observation stands out. Specifically ReliefF 100 NN, fails in all configurations where $n=200$. For a $k$ of 100 in a balanced dataset (such as this), ReliefF is using all other instances as neighbors. As we reviewed by \citet{robnik2003comprehensible}, this effectively removes the requirement that instances used in scoring be `near' and turns ReliefF (using all neighbors) into a myopic algorithm, unable to handle 2-way interactions. This is verified empirically by these results. In contrast, examination of ReliefF 10 NN results (e.g. heritability = 0.05, $n=800$ or $n=1600$, difficulty = E) in contrast to other RBAs suggests that increasing noise (i.e. lower heritability) is better handled by a somewhat larger $k$. This is consistent with previous observations \citep{kononenko1994estimating,greene2009spatially}. It should be noted the concept of a `low' or `high' $k$ should always be considered with respect to $n$. For example, while $k=100$ was detrimental in a sample size of 200, this setting performed quite well when $n$ was 400 to 1600.  

Regarding the the ExtraTrees wrapper algorithms, we note that these random forest approaches were able to detect pure 2-way epistatic interactions (at least in datasets with 20 features). Comparing them to RBAs we observe random forest performance to be more negatively impacted by increased noise and decreased $n$.

Focusing on SURF, SURF*, and MultiSURF*, the results in Figure \ref{fig:Results:2-way} subtly but consistently support the previous findings that over a spectrum of 2-way pure epistatic interactions, SURF $<$ SURF* $<$ MultiSURF* with respect to power \citep{greene2009spatially,greene2010informative,granizo2013multiple}. Lastly, examination of our proposed MultiSURF variant, suggests that it's performance is generally competitive with other RBAs (on par with SURF \citep{greene2009spatially}), but is slightly outperformed by SURF* and MultiSURF* (both methods that adopt far scoring) on 2-way epistasis problems with increased noise and decreased $n$. For a detailed discussion regarding why far scoring improves the detection of 2-way interactions, see Section \ref{discussion}.

\subsection{Main Effects}

\begin{figure}[t]
	\centerline{\includegraphics[width=\textwidth]{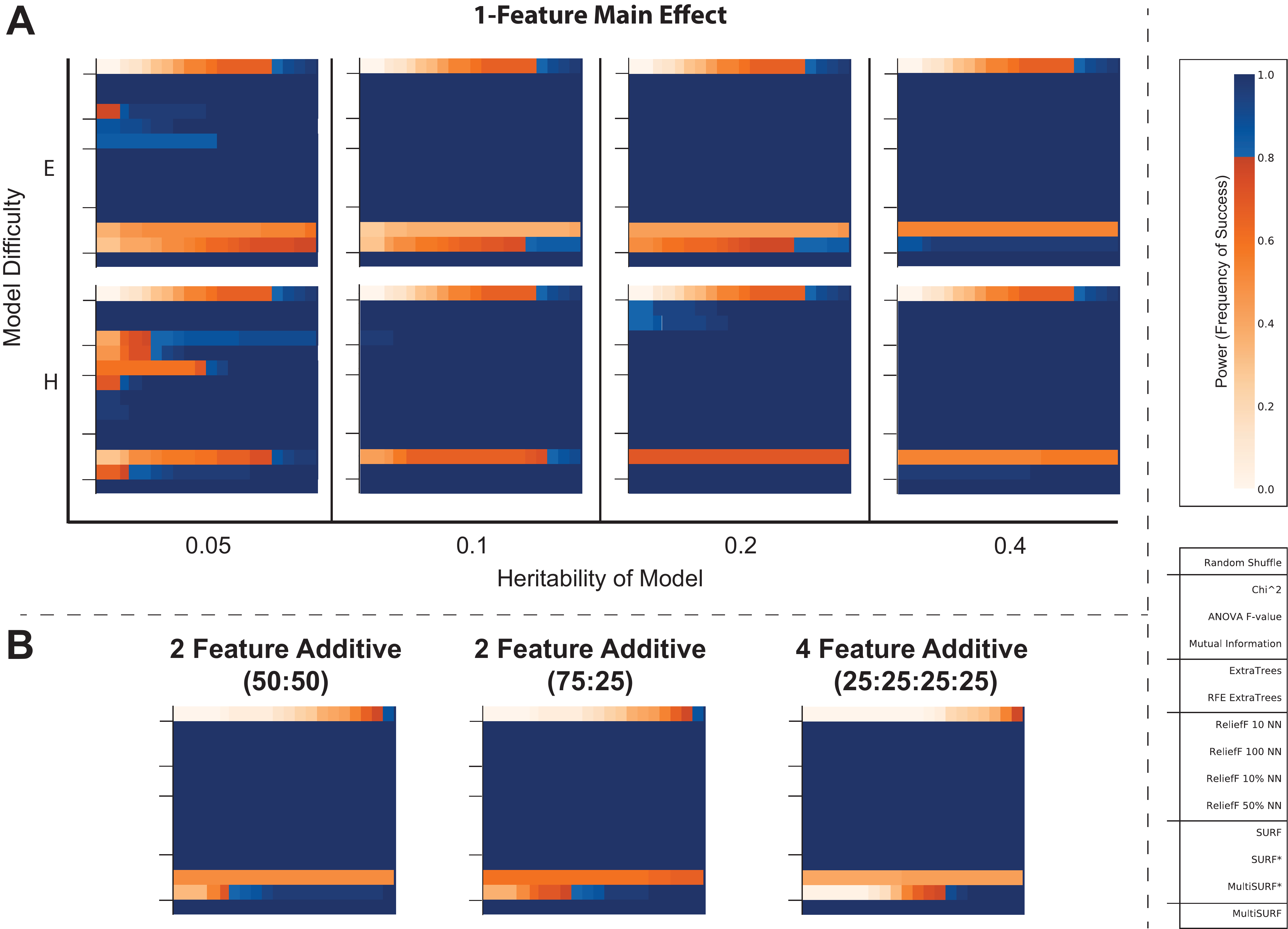}}
	\caption{Results for detecting single feature main effects (A) and additive main effects (B). Keys relevant to all plots are given on the far right. Tick marks delineating algorithm groups are provided for each sub-plot.}
	\label{fig:Results:Main}
\end{figure}

Main effects (i.e. the effect of a single independent variable on a dependent variable) are generally understood to be easier and less computationally expensive to detect than interactions. It is likely for this reason that the RBAs in ReBATE had never been tested on simulations of single feature main effects or multiple feature additive main effects. Figure \ref{fig:Results:Main}A presents main effect results over increasing heritabilities (x-axis) and model difficulty (y-axis). First, notice that all three myopic filter algorithms (i.e. chi squared test, ANOVA F-value, and mutual information) generally succeed at identifying the respective main effects. Of the three, mutual information was least successful, particularly when heritability was low and the model was `hard'. A similar performance loss was observed in the random forest wrappers.

The most dramatic and unexpected finding in this analysis was the performance loss of SURF* and MultiSURF*, the only two algorithms utilizing `far' scoring. Interestingly, this loss was even more significant in easy main effect models. This is in contrast with ReliefF, SURF, and our proposed MultiSURF algorithm that were completely successful here. Examination of additive multi-feature main effects in Figure \ref{fig:Results:Main}B each with a heritability of 0.4 and `E' model, reveal similar performance losses for SURF* and MultiSURF*. Notably, for all main effect datasets, MultiSURF* has less of a performance loss than SURF*. For a detailed discussion regarding why far scoring hinders or eliminates the ability to detect main effects, see Section \ref{discussion}.

\subsection{Genetic Heterogeneity}

\begin{figure}[t]
	\centerline{\includegraphics[width=\textwidth]{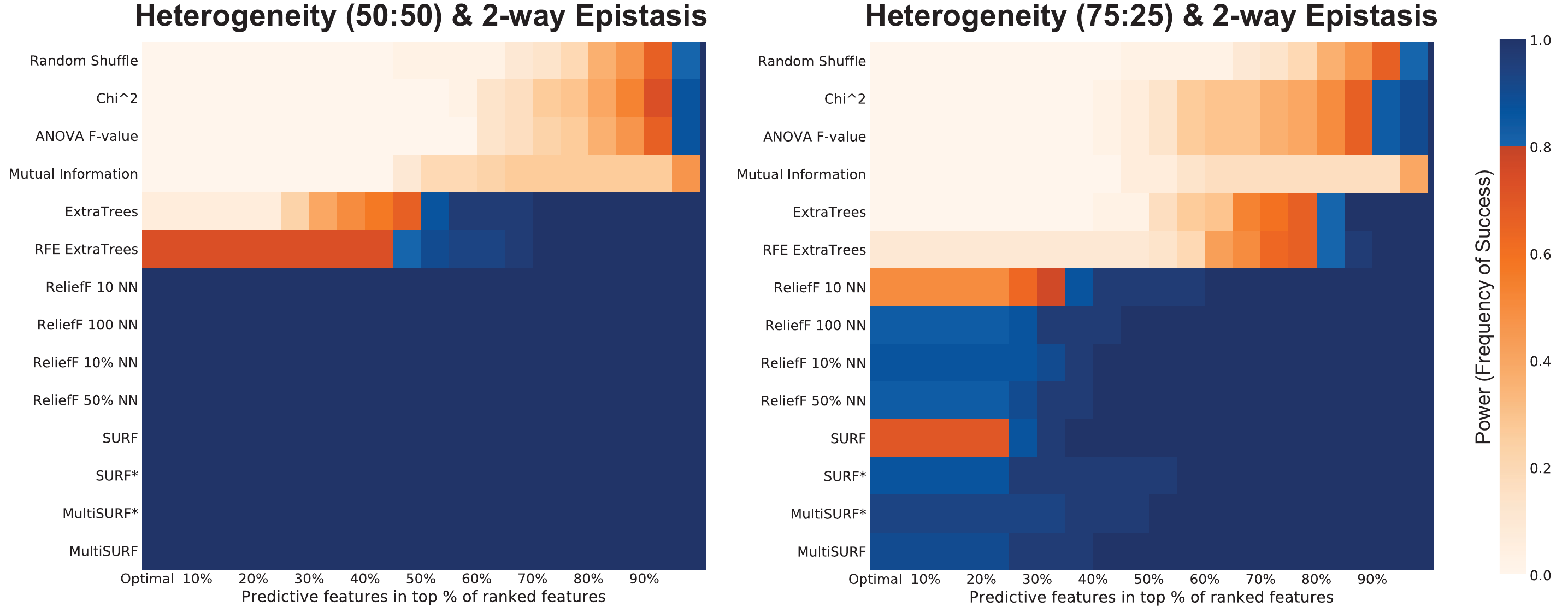}}
	\caption{Results for detecting two independent heterogeneous 2-way epistatic interactions.}
	\label{fig:Results:Het}
\end{figure}

Figure \ref{fig:Results:Het} gives the power analysis results for data that models heterogeneous patterns of association between independent 2-way epistatic interactions. In other words, in one subset of training instances one 2-way interaction is relevant, and in the other a different pair of interacting features are relevant. The given ratio indicates the proportion of instances within which each interaction is relevant. Heterogeneous patterns of association have been commonly recognized in biomedical problems and are known to confound traditional machine learning approaches \citep{ritchie2003power,thornton2006dissecting,urbanowicz2010application,urbanowicz2013role}. Accounting for such patterns in feature selection is thus an important target. Figure \ref{fig:Results:Het} suggests that all tested RBAs can handle heterogeneity concurrently modeled with epistatic interactions, while all other methods fail, or fail to perform nearly as well, in the case of the random forest wrappers. With a more extreme ratio of 75:25, MultiSURF*, and MultiSURF appear to perform best with SURF* and ReliefF (with larger $k$ settings) close behind. Overall, RBAs in general appear uniquely suited to detecting patterns of both heterogeneous association. To the best of our knowledge this is the first formal evaluation of RBAs on heterogeneous patterns of association.

\subsection{3-way Epistasis}
Referring back to Figure \ref{fig:Results:3-way} we provide the first evaluation of RBA performance on epistatic interactions with a dimmensionality higher than 2 (i.e. 3-way interactions). Due to mathematical constraints, the GAMETES epistasis dataset simulation software was unable to generate a 3-way SNP interaction dataset with a heritability of 0.4, so instead we simulated 3-way interaction SNP datasets with heritability=0.2. As expected, the myopic methods fail to perform well. This is also true for the ExtraTrees wrappers.  Interestingly, the only RBAs that succeeded on this problem were those that utilized the smallest number of neighbors in scoring (i.e. ReliefF with 10 or 100 NN, ReliefF with $10\%$ NN, and our proposed MultiSURF algorithm). This suggests that detecting higher order interactions is best achieved when the number of neighbors is low with respect to $n$.  In this analysis the dataset included 1600 instances. ReliefF performed well with 10, 100, or 80 ($0.1*1600/2$) nearest hits and misses, but completely failed with 400 ($0.5*1600/2$) nearest hits and misses. We expect that MultiSURF $<$ MultiSURF* $<$ SURF $<$ SURF* with respect to the number of instances involved in each scoring cycle, where SURF* utilizes all instances in scoring and MultiSURF should utilize the least given that scoring only includes the inside of the dead-band zone.

There is no evidence to suggest that `far' scoring itself is helpful or harmful in this analysis. This is one prominent scenario where our proposed MultiSURF algorithm succeeds where other modern RBAs fail. While ReliefF performs slightly better, it's success is dependent on setting $k$ properly. It may be useful to track the number of instances involved in scoring in these algorithm in future investigations of RBAs detecting higher order interactions. 

\subsection{Number of Features}

\begin{figure}[t]
	\centerline{\includegraphics[width=\textwidth]{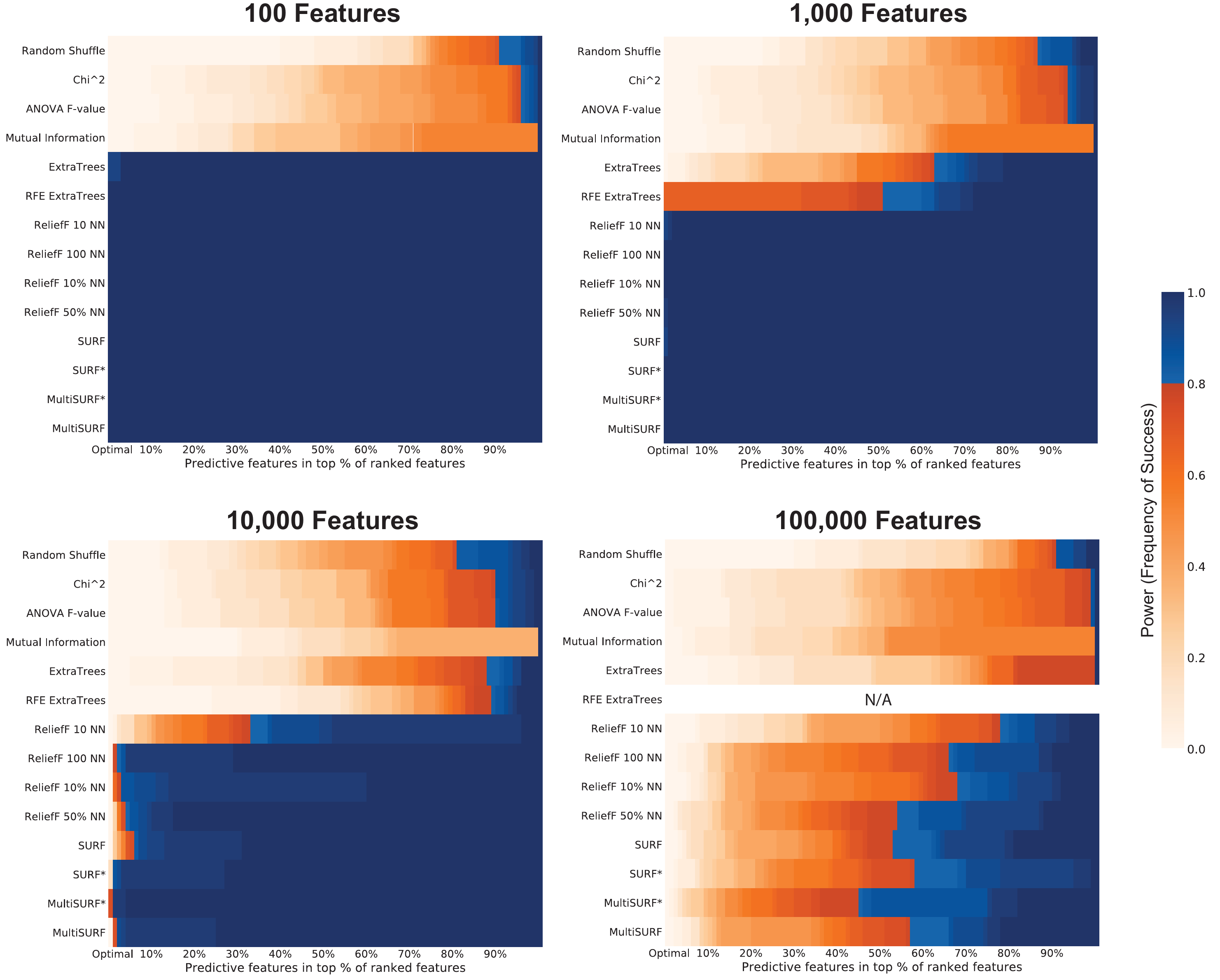}}
	\caption{Results for detecting 2-way epistatic interactions with an increasing number of irrelevant features in the datasets.}
	\label{fig:Results:Features}
\end{figure}

All of the results presented so far were run on datasets with a relatively small number of features to save computational time while asking basic questions about algorithm abilities. Figure \ref{fig:Results:Features} presents algorithm power over an increasing number of irrelevant features in datasets with 2-way epistatic interactions. We examined feature space sizes up to the maximum number of features investigated in a previous simulation study of an RBA (i.e 100,000) \citep{eppstein2008very}. Keep in mind that we don't expect any of these core methods to perform particularly well in very large feature spaces without ultimately combining them with some iterative RBA approach. As usual, the myopic approaches fail to detect 2-way interactions. The next methods to fail in a feature space of increasing size are the random forest wrappers (at 1000 features). At 10,000 features, ReliefF with $k=10$ begins to fail suggesting that a small number of neighbors performs less well in noisy problems, particularly as the feature space grows. At 10,000 features, MultiSURF* appears to perform slightly better than the rest, consistent with our previous findings in Figure \ref{fig:Results:2-way}. Lastly, as expected at 100,000 features none of the methods perform particularly well on their own, but still certainly better than a random shuffle. Results for RFE ExtraTrees are missing because this iterative random forest approach did not finish running within a reasonable amount of time (i.e. over two days). Based on the results of RFE ExtraTrees in smaller feature sets, it is reasonable to assume it would have performed poorly at 100,000 features as well. Notably, most RBAs (with the exception of ReliefF 10 NN, 100 NN, and $10\%$ NN) demonstrate high power (i.e. $>80\%$) to rank the predictive features above the 60th percentile, and MultiSURF* is the only method with significant power above the 50th percentile. Also of note, if we look for the percentile at which algorithms acheive full power, our experimental MultiSURF achieves the best percentile (i.e. approximately the 80th percentile). In combination with an iterative approach, for example TuRF, the lowest ranking features would be removed each iteration, improving the estimation of remaining feature scores in subsequent iterations. Therefore it is useful here to consider what proportion of underlying features could be removed in the first (and subsequent) iterations without losing any relevant features. 

\subsection{XOR Benchmarks}

\begin{figure}[t]
	\centerline{\includegraphics[width=\textwidth]{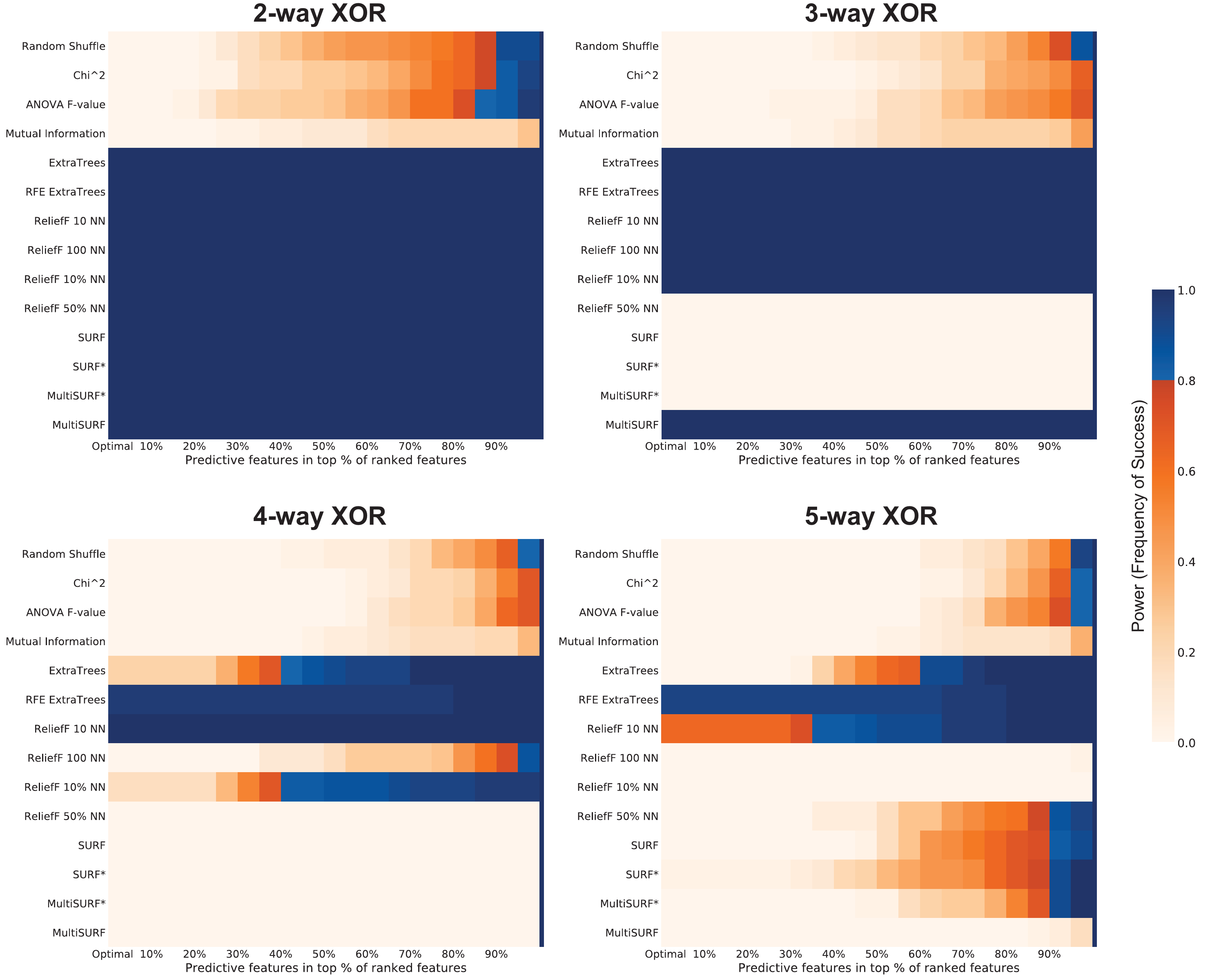}}
	\caption{Results for detecting 2-way, 3-way, 4-way, and 5-way epistatic interactions based on `clean' XOR models.  }
	\label{fig:Results:XOR}
\end{figure}

Here we examine clean benchmark datasets that further test epistatic patterns. Figure \ref{fig:Results:XOR} presents power analyses for increasingly higher order versions of the XOR problem simulated here as SNP data \citep{urbanowicz2012gametes}. These XOR benchmark datasets have no noise and include pure epistatic interactions from 2-way up to 5-way interactions. While all non-myopic methods solve the 2-way XOR with little trouble, we observe a similar pattern of algorithm success for the 3-way XOR as we did with the noisy GAMETES generated 3-way dataset in Figure \ref{fig:Results:3-way}. Specifically, RBAs that utilized fewer neighbors were successful, including ReliefF with 10 or 100 NN, ReliefF with $10\%$ NN and our proposed MultiSURF algorithm. Differently, the ExtraTrees algorithms were equally successful in detecting the 3-way XOR as those RBAs. Another interesting observation for the 3-way XOR is that for RBAs that used a larger proportion of neighbors in scoring, relevant features were consistently ranked with the lowest overall scores. This was also true for some of the 4-way and 5-way analyses. This is interesting because if the methods were not detecting any difference between relevant and irrelevant features, we would expect to see results similar to the random shuffle negative control. Instead, in specific high-order interaction problems, relevant features are consistently being assigned the most negative score updates. There may be opportunity for future RBA improvement leveraging this observation. Currently this is a problem for those respective RBAs since users will assume the at features scoring at the bottom of the feature ranking should be eliminated from consideration. It is only because we know the ground truth of these simulated datsets that we can observe this pattern here.

Overall, the only method able to solve all XOR problems is the RFE ExtraTrees, however based on our previous findings we expect this success to quickly disappear if the feature space was larger than 20 features. The next best algorithm is ReliefF with 10 neighbors, followed by ExtraTrees and ReliefF with $10\%$ (i.e 80 neighbors). Notably, the the 4-way and 5-way XOR problems represent one situation in which our proposed MultiSURF algorithm fails to perform. It is clear that higher order interactions can be problematic for RBAs unless they apply few neighbors in scoring. 

\subsection{Multiplexer Benchmarks}

\begin{figure}[t]
	\centerline{\includegraphics[width=\textwidth]{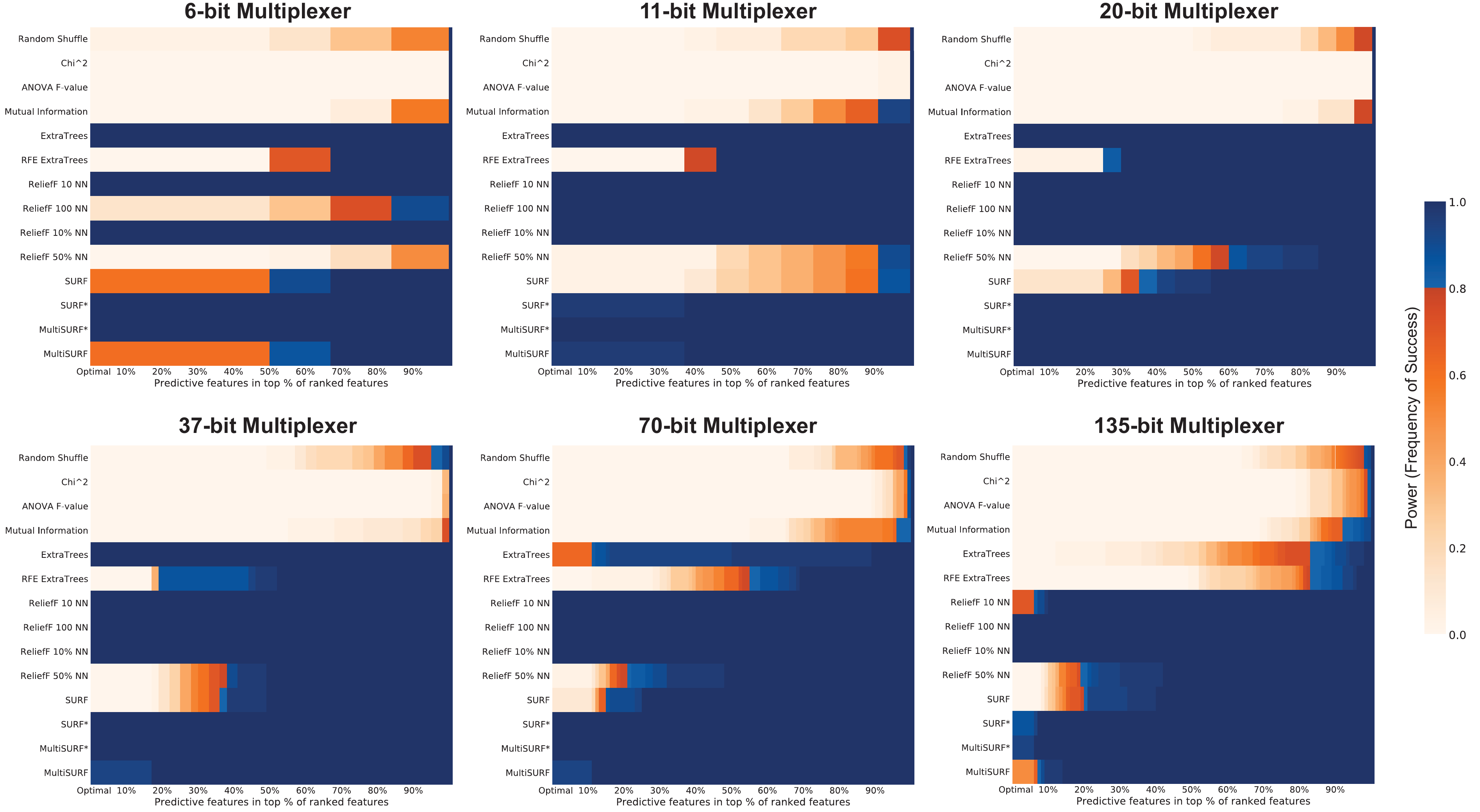}}
	\caption{Results for detecting the address bits of different scalings of the Mulitplexer benchmark problem.  Each problem is `clean', epistatic, and heterogeneous.  Note that all features in these datasets are predictive in at least one subset of the training instances, and power reflects the ability to rank the subset of features that are important in all training instances (address bits), from those that are important only in a given subset (register bits). }
	\label{fig:Results:MUX}
\end{figure}

Figure \ref{fig:Results:MUX} presents power analyses for increasingly higher order versions of the multiplexer benchmark datasets including the 6-bit through the 135-bit versions. This power analysis should be interpreted somewhat differently from all others since in these datasets, technically all features are relevant (in at least some subset of training instances), and instead we evaluate the power to properly rank the address bits of the multiplexer problem (i.e. the features that are relevant for every instance in the dataset.  We had previously observed in \citep{urbanowicz2015exstracs} that MultiSURF*'s ability to rank address bits over all others in feature weighting was one of the keys to solving the 135-bit multiplexer benchmark directly for the first time in the litterature. 

Like the XOR problem, the multiplexer problems have no main effects. As a result, the myopic methods again fail to perform on these datasets. The only method to perform perfectly over all datasets was Relief $10\%$ NN. MultiSURF* perfromed next best followed by SURF* and our MultiSURF method trailing close behind. These results are interesting but difficult to clearly interpret given that all features are technically predictive (particularly with respect to the 6-bit problem which is the `easiest' to solve). However, this explains why feature weighting using MultiSURF* in ExSTraCS was facilitated solving the 135-bit multiplexer. These results also emphasize the ability of RBAs in general to detect relevant features in the presence of both feature interactions and heterogeneous patterns of association.

\subsection{Data Types}

\begin{figure}[t]
	\centerline{\includegraphics[width=\textwidth]{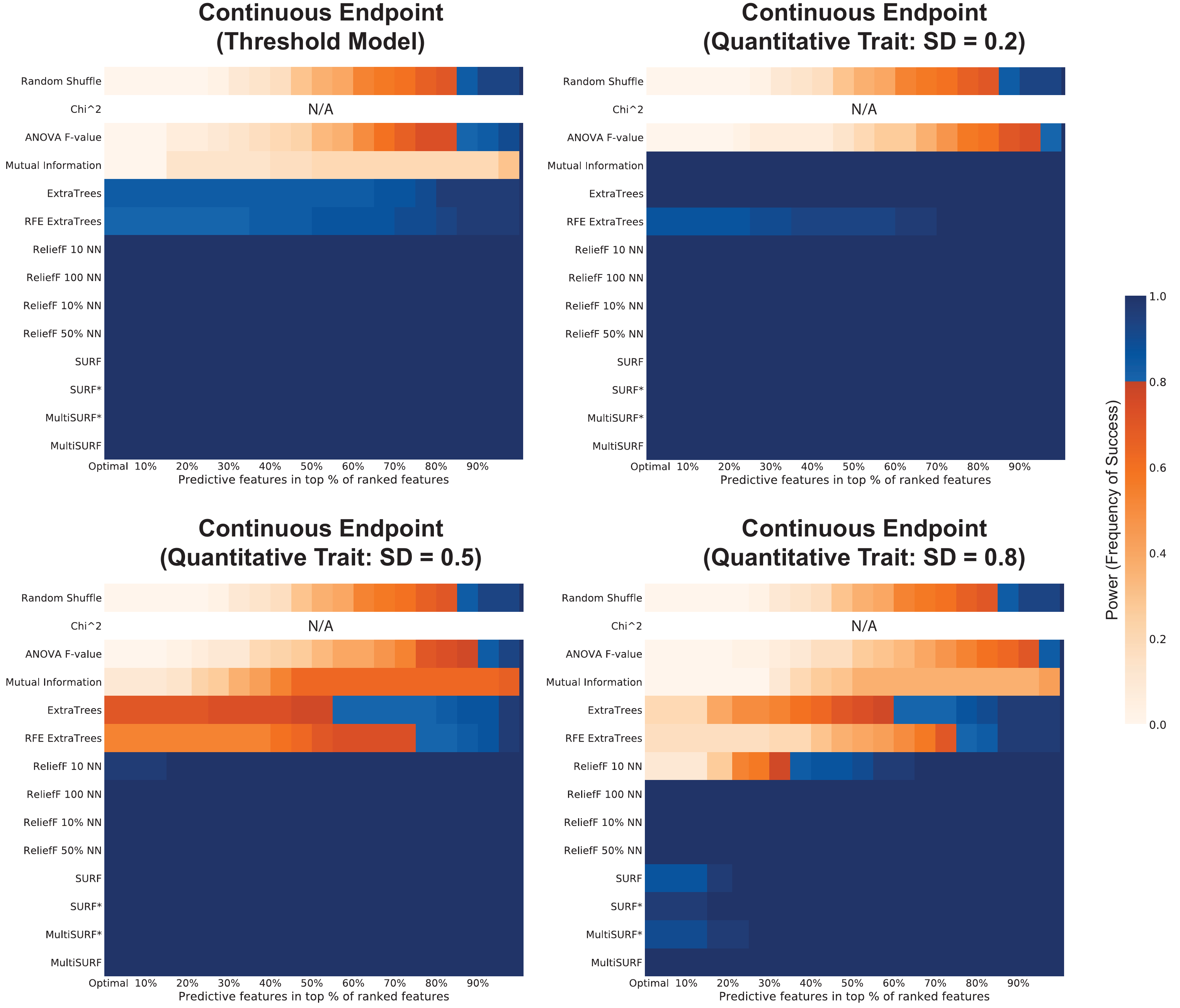}}
	\caption{Results for accommodating continuous (i.e. numerical) endpoints in datasets.}
	\label{fig:Results:CE}
\end{figure}

We conclude our results section with power analyses for the different data type scenarios considered. This will demonstrate the functionality of our ReBATE algorithm data type extensions. Only power analyses yielding interesting differences are explicitly presented. 

Figure \ref{fig:Results:CE} gives power results over four numerical endpoints configurations, each simulated with an underlying 2-way interaction. First, note the the Chi Square test is not applicable to data with continuous endpoints. As usual, the other myopic methods did not perform well with the exception of mutual information (on continuous endpoint data generated with a standard deviation of 0.2). It is possible that the strategy we employed to generate continuous endpoint SNP data from epistatic models introduced some small main effects at a low standard deviation that mutual information was able to pick up. Interestingly, the random forest methods performed slightly worse than the RBAs. Overall, the success of all RBAs implemented in ReBATE in these analyses suggests that our simpler and computationally less expensive proposed RBA regression approach offers a functional alternative to the one proposed in RReliefF \citep{kononenko1996relieff,robnik1997adaptation}. A comparison between the two is outside the scope of the current investigation but should be the focus of future study.

\begin{figure}[t]
	\centerline{\includegraphics[width=\textwidth]{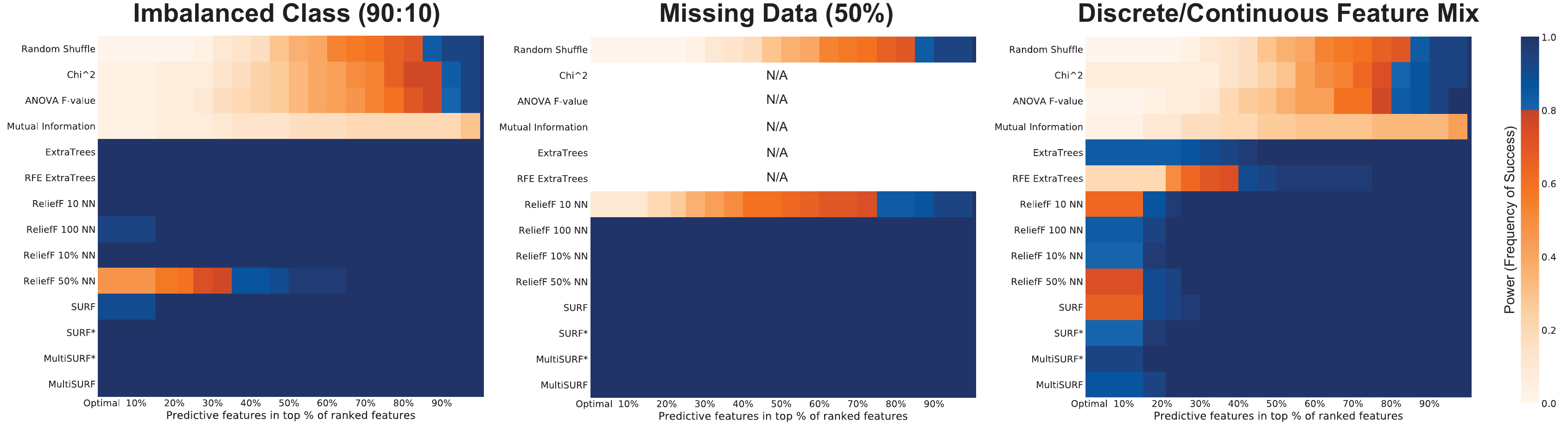}}
	\caption{Results for accommodating different `data type' issues. Specifically this figure examines extreme examples of class imbalance, missing data, and the combination of discrete and continuous features within the same dataset.}
	\label{fig:Results:DT}
\end{figure}

Most of our power analyses for the other data type configurations yielded the same successful results for all RBAs, thus results are not presented here. Specifically, both multi-class data configurations were solved by all feature selection methods, demonstrating the basic efficacy of the multi-class expansion for Relief introduced in \citep{kononenko1994estimating} and adopted in ReBATE. This also indicated that our simulated multi-class datasets (with impure epistasis) had strong enough main effects to be successfully ranked by all myopic methods. 

For datasets with only continuous features and an underlying 2-way interaction, all but the myopic methods ranked features ideally, reinforcing the functionality of the proposed \emph{diff} function for continuous features originally introduced in \citep{kira1992practical}. The same was true for data with an imbalance of 0.6 (i.e. $60\%$ class 0, $40\%$ class 1), and for missing data with data randomly missing at a frequency of either 0.001, 0.01, or 0.1. Notably in the analyses of missing data, all 5 feature selection methods run with scikit-learn (i.e. chi square test, ANOVA F-test, mutual information, ExtraTrees, and RFE ExtraTrees) could not be completed since scikit-learn is not set up to handle missing data. Preprocessing such as removal of instances with missing values or imputation would instead be required. 

Figure \ref{fig:Results:DT} presents all remaining results in which feature selection performance differences were observed. First off, for each, myopic approaches again failed to detect the underlying 2-way interactions. For a class imbalance of 0.9 (i.e. $90\%$ class 0, $10\%$ class 1), we observe that ReliefF with a large number of neighbors (i.e. $50\%$) fails to perform, ReliefF with 100 NN and SURF demonstrate slight deficits, but all other RBAs perform optimally. For missing data with a frequency of 0.5, we observe all ReliefF methods perform optimally with the exception of ReliefF 10 NN (the fewest neighbors in scoring). This suggests that having more neighbors in scoring makes these methods more resiliant to missing data, and also demonstrates that our proposed agnostic missing data strategy implemented in ReBATE is successful.  A comparison between different missing data handling strategies is beyond the scope of this investigation, but should be examined in future study. Lastly, Figure \ref{fig:Results:DT} gives the results for the data configuration with a mix of discrete and continuous features. For each of the 30 dataset replicates a random half of the features is re-encoded with a continuous value, and the rest are left as discrete values. Similar to previous observations regarding mixed feature types with Relief \citep{kononenko2008non}, none of the RBAs or feature selection methods in general handle this dataset configuration optimally. MultiSURF* works best followed by MultiSURF, SURF*, and ReliefF with 100 NN or $10\%$ NN. Overall this suggests that mixed feature types are still an issue for RBAs, and worth further study and development in ReBATE.  Future work should consider the proposed ramp function \citep{hong1997use,robnik2003theoretical,kononenko2008non}, as well as other approaches that do not require setting user parameters to address this issue. 

\section{Discussion} \label{discussion}
In this section we discuss why specific strengths and weaknesses were observed for respective methods under different data configurations.

\subsection{2-way Epistastis Performance Gains}
From the results, one may wonder why is it that `far' scoring in SURF* and MultiSURF* appears to enhance power to detect 2-way interactions? To help answer this question, Table \ref{tab:Epi}, offers a simple example 2-way epistasis dataset that we will use to walk through Relief scoring. In this example, $A_{1}$ and $A_{2}$ are relevant features with a pure interaction between them (i.e. no individual main effects). When they have opposing values, the class is one otherwise the class is zero. $A_{3}$ is an irrelevant feature.  

Table \ref{tab:Epi2} breaks down how scoring would proceed over 8 cycles with each instance getting to be the respective target. To simplify this example, we only select one near or far instance for each RBA, focusing instead on the impact of scoring scheme. Notice that we will keep track of nearest hits and misses (and the corresponding features with a different value), as well as far instances used in scoring by SURF* and MultiSURF* (i.e. the farthest hit or miss, each with both the same and different feature value(s) identified between instances). For each target, we see what instance is the nearest or farthest hit and miss, as well as which feature has a different value between the instances (given in parentheses). Further, we see what feature has the same value between farthest instances (in grey). All of these will be relevant to scoring in at least one of the ReBATE algorithms. If there is a tie for nearest neighbor, both instances are listed with their respective different valued feature. For example, when $R_{1}$ is the target, it's nearest hit is $R_{2}$. The only feature with a different value between these two instances is $A_{3}$. The nearest miss for $R_{1}$ is a tie between $R_{5}$ and $R_{7}$ that have feature value differences at $A_{1}$ and $A_{2}$, respectively. The farthest miss for $R_{1}$ is a tie between $R_{6}$ and $R_{8}$ that have feature value differences at $A_{1}$ and $A_{3}$, and $A_{2}$ and $A_{3}$, respectively. Lastly, we keep track of the feature values that are the same between the farthest hits and misses (grey shaded cells).  For the farthest hit, no feature values are the same since we already noted they were all different. For the farthest miss, we again have $R_{6}$ and $R_{8}$ that have feature value equalities at $A_{2}$ and $A_{1}$, respectively.

\begin{table*}[h]
 \centering
\caption{Example dataset with interaction between $A_{1}$ and $A_{2}$. $A_{3}$ is irrelevant. Adapted from \citet{kononenko1997overcoming}.}
\label{tab:Epi}       
{\small 
\begin{tabular}{|c||ccc|c|}
\hline
Instances & $A_{1}$ & $A_{2}$ & $A_{3}$ & $C$ \\ \hline
$R_{1}$ & 1 & 0 & 1 & 1   \\
$R_{2}$ & 1 & 0 & 0 & 1   \\
$R_{3}$ & 0 & 1 & 1 & 1   \\
$R_{4}$ & 0 & 1 & 0 & 1   \\
$R_{5}$ & 0 & 0 & 1 & 0   \\
$R_{6}$ & 0 & 0 & 0 & 0   \\
$R_{7}$ & 1 & 1 & 1 & 0   \\
$R_{8}$ & 1 & 1 & 0 & 0   \\

\hline
\end{tabular}
}
\end{table*}

\begin{table*}[h]
 \centering
\caption{Breakdown of near/far, hit/miss, and features with different or same values for each target instance. Illustrates where feature scores come from for the 2-way epistasis dataset in Table \ref{tab:Epi}.}
\label{tab:Epi2}       
{\footnotesize 
\begin{tabular}{|c|c|c|c|c|c|c|}
\cline{2-7}
\multicolumn{1}{c|}{} & \multicolumn{2}{c|}{\cellcolor{gray!30}Far (Same Value)} & \multicolumn{2}{c|}{Near (Different Value)} & \multicolumn{2}{c|}{Far (Different Value)} \\ \hline
\multicolumn{1}{|c|}{Target} & \multicolumn{1}{c|}{\cellcolor{gray!30}Hit} & \multicolumn{1}{c|}{\cellcolor{gray!30}Miss} & \multicolumn{1}{c|}{Hit} & \multicolumn{1}{c|}{Miss} & \multicolumn{1}{c|}{Hit} & \multicolumn{1}{c|}{Miss}  \\ \hline

$R_{1}$&\cellcolor{gray!30}None&\cellcolor{gray!30}$R_{6}(A_{2})$,$R_{8}(A_{1})$& $R_{2}(A_{3})$ & $R_{5}(A_{1})$,$R_{7}(A_{2})$ & $R_{4}(A_{1},A_{2},A_{3})$ &$R_{6}(A_{1},A_{3})$,$R_{8}(A_{2},A_{3})$  \\
$R_{2}$&\cellcolor{gray!30}None&\cellcolor{gray!30}$R_{5}(A_{2})$,$R_{7}(A_{1})$& $R_{1}(A_{3})$ & $R_{6}(A_{1})$,$R_{8}(A_{2})$ & $R_{3}(A_{1},A_{2},A_{3})$ &$R_{5}(A_{1},A_{3})$,$R_{7}(A_{2},A_{3})$  \\
$R_{3}$&\cellcolor{gray!30}None&\cellcolor{gray!30}$R_{6}(A_{1})$,$R_{8}(A_{2})$& $R_{4}(A_{3})$ & $R_{5}(A_{2})$,$R_{7}(A_{1})$  & $R_{2}(A_{1},A_{2},A_{3})$ &$R_{6}(A_{2},A_{3})$,$R_{8}(A_{1},A_{3})$ \\
$R_{4}$&\cellcolor{gray!30}None&\cellcolor{gray!30}$R_{5}(A_{1})$,$R_{7}(A_{2})$& $R_{3}(A_{3})$ & $R_{6}(A_{2})$,$R_{8}(A_{1})$ & $R_{1}(A_{1},A_{2},A_{3})$ &$R_{5}(A_{2},A_{3})$,$R_{7}(A_{1},A_{3})$  \\
$R_{5}$&\cellcolor{gray!30}None&\cellcolor{gray!30}$R_{2}(A_{2})$,$R_{4}(A_{1})$ & $R_{6}(A_{3})$ & $R_{1}(A_{1})$,$R_{3}(A_{2})$ & $R_{8}(A_{1},A_{2},A_{3})$ &$R_{1}(A_{1},A_{3})$,$R_{4}(A_{2},A_{3})$ \\
$R_{6}$&\cellcolor{gray!30}None&\cellcolor{gray!30}$R_{1}(A_{2})$,$R_{3}(A_{1})$& $R_{5}(A_{3})$ & $R_{2}(A_{1})$,$R_{4}(A_{2})$  & $R_{7}(A_{1},A_{2},A_{3})$ &$R_{1}(A_{1},A_{3})$,$R_{3}(A_{2},A_{3})$ \\
$R_{7}$&\cellcolor{gray!30}None&\cellcolor{gray!30}$R_{2}(A_{1})$,$R_{4}(A_{2})$ & $R_{8}(A_{3})$ & $R_{1}(A_{2})$,$R_{3}(A_{1})$ & $R_{6}(A_{1},A_{2},A_{3})$ &$R_{2}(A_{2},A_{3})$,$R_{4}(A_{1},A_{3})$ \\
$R_{8}$&\cellcolor{gray!30}None&\cellcolor{gray!30}$R_{1}(A_{1})$,$R_{3}(A_{2})$& $R_{7}(A_{3})$ & $R_{2}(A_{2})$,$R_{4}(A_{1})$ & $R_{5}(A_{1},A_{2},A_{3})$ &$R_{1}(A_{2},A_{3})$,$R_{3}(A_{1},A_{3})$  \\
\hline
\end{tabular}
}
\end{table*}

\begin{table*}[h]
 \centering
\caption{Summary of score contributions for 2-way epistasis dataset in Table \ref{tab:Epi} yielding ReliefF, SURF, SURF*, MultiSURF*, and MultiSURF scores. Illustrates why `far' scoring improves 2-way interaction detection.}
\label{tab:Episum2}       
{\small 
\begin{tabular}{|c||c|c|c||c|c|c|c|c|}

\cline{2-9}
\multicolumn{1}{c||}{} & \multicolumn{3}{c||}{Features} & \multicolumn{5}{c|}{RBA Scoring Schemes}  \\ \cline{2-9}
\multicolumn{1}{c||}{}& & & & \multirow{6}{*}{\rotatebox{90}{ReliefF}} & \multirow{6}{*}{\rotatebox{90}{SURF}} & \multirow{6}{*}{\rotatebox{90}{SURF*}} & \multirow{6}{*}{\rotatebox{90}{MultiSURF*}} & \multirow{6}{*}{\rotatebox{90}{MultiSURF}}  \\ 
\multicolumn{1}{c||}{}& & & & & & & &   \\ 
\multicolumn{1}{c||}{}& & & & & & & &   \\ 
\multicolumn{1}{c||}{}& & & & & & & &   \\ 
\multicolumn{1}{c||}{}& & & & & & & &   \\ 
\multicolumn{1}{c||}{}& $A_{1}$&$A_{2}$ & $A_{3}$& & & & &  \\  \hline \hline

\rowcolor{gray!30}Farthest Hit (same value)& 0 & 0 & 0  && && - 1 & \\
\rowcolor{gray!30}Farthest Miss (same value)& 4 & 4 & 0 && && +1 & \\ \hline 
Nearest Hit (different value)& 0 & 0 & 8 & - 1  &- 1 &- 1&- 1&- 1  \\
Nearest Miss (different value)& 4 & 4 & 0 & +1 &+1 &+1&+1&+1  \\ \hline 
Farthest Hit (different value)& 8 & 8 & 8 & &   &+1 && \\
Farthest Miss (different value)& 4 & 4 & 8 & &   &- 1&& \\ \hline 
\multicolumn{9}{c}{}\\ \cline{1-4}

\multicolumn{1}{|c||}{ReliefF Score Total} & \multicolumn{1}{c|}{\cellcolor{green!15}4} & \multicolumn{1}{c|}{\cellcolor{green!15}4} & \multicolumn{1}{c||}{\cellcolor{red!30}-8} & \multicolumn{5}{c}{} \\ \cline{1-4}
\multicolumn{1}{|c||}{SURF Score Total} & \multicolumn{1}{c|}{\cellcolor{green!15}4} & \multicolumn{1}{c|}{\cellcolor{green!15}4} & \multicolumn{1}{c||}{\cellcolor{red!30}-8} & \multicolumn{5}{c}{} \\ \cline{1-4}
\multicolumn{1}{|c||}{SURF* Score Total} & \multicolumn{1}{c|}{\cellcolor{green!30}8} & \multicolumn{1}{c|}{\cellcolor{green!30}8} & \multicolumn{1}{c||}{\cellcolor{red!30}-8} & \multicolumn{5}{c}{} \\ \cline{1-4}
\multicolumn{1}{|c||}{MultiSURF* Score Total} & \multicolumn{1}{c|}{\cellcolor{green!30}8} & \multicolumn{1}{c|}{\cellcolor{green!30}8} & \multicolumn{1}{c||}{\cellcolor{red!30}-8} & \multicolumn{5}{c}{} \\ \cline{1-4}
\multicolumn{1}{|c||}{MultiSURF Score Total} & \multicolumn{1}{c|}{\cellcolor{green!15}4} & \multicolumn{1}{c|}{\cellcolor{green!15}4} & \multicolumn{1}{c||}{\cellcolor{red!30}-8} & \multicolumn{5}{c}{} \\ 
\cline{1-4}
\end{tabular}
}
\end{table*}

Table \ref{tab:Episum2} summarizes the resulting score contributions from Table \ref{tab:Epi2}. As before, when there is a tie between instances for nearest or farthest neighbor, we give each feature difference half credit since only one can contribute at a time. For example, since there are no same feature values for any of the farthest hits in Table \ref{tab:Epi2} (shaded in grey), all three features get a zero score contribution under `Farthest Hit (same value)' in Table \ref{tab:Episum2}. 

As Table \ref{tab:Episum2} depicts, all five scoring approaches assign positive scores to $A_{1}$ and $A_{2}$, and a negative score of $-8$ to $A_{3}$. Thus, all of these methods can differentiating relevant from irrelevant features in the presence of a pure 2-way interaction. However `far' scoring in SURF* and MultiSURF* make the score difference between a relevant and irrelevant feature score even greater. In the context of 2-way interactions, `far' scoring serves to reinforce scores of interacting features by offering a larger `scoring sample size', as originally intended in \citep{greene2010informative}. However as this example breakdown also indicates, the scoring approach does not explain performance differences between SURF* and MultiSURF* in detecting 2-way interactions.  That is likely the result of MultiSURF*'s target-centric threshold calculation and/or its application of a dead-band zone.

\subsection{Main Effects Performance Losses}
To understand why `far' scoring in SURF* and MultiSURF* leads to main effect performance loss, we lay out a conceptual scenario in Table \ref{tab:main}. This scenario assumes an example problem with binary features and endpoint. On the far left we have differently labeled instances including a hypothetical target instance ($R_{i}$), as well as every other possible type of instance from the perspective of the target instance $R_{i}$. Other instance types are differentiated based on properties such as `Distance', (near or far), `Class' (hits or misses), and if the relevant feature (i.e. with a main effect) has the same or different value as $R_{i}$ (instance types with the same feature value are shaded in grey). For simplicity we will assume that all irrelevant features are expected to end up with a feature score of approximately zero as derived in \citep{kira1992feature}.

The key to understanding this conceptual scenario is understanding where the expected frequencies of each instance type come from. To do that, first it's important to define a few expectations of datasets that include a main effect: (1) irrelevant feature values will be randomly distributed, (2) instances that have a relevant feature with the same value between them will tend to be closer on average (assuming the first expectation), and (3) if at least one feature in the dataset is relevant, we expect instances with the same class to be closer on average (since the one relevant feature will typically have the same feature value between instances with the same class value). Based on these expectations we can describe `likely' combinations of distance with class, and distance with feature value. These include the following; (1) near and same class, (2) near and same feature value, (3) far and different class, and (4) far and different feature value. All other combinations are less likely based on our expectations. Finally we estimate the frequency of an instance `type' based on the number of `likely' combinations it includes. Specifically, if it has two likely combinations it is `high' frequency, if it has two unlikely combinations it is `low' frequency, and if it has one of each it has `medium' frequency.

\begin{table*}[h!]
 \centering
\caption{Conceptual scenario illustrating why `far' scoring deteriorates main effect performance.}
\label{tab:main}       
{\scriptsize 
\begin{tabular}{|c||c|c|c|c|c||c|c|c|c|c|}
\cline{2-11}
\multicolumn{1}{c||}{} & \multicolumn{5}{c||}{Instance `Type' Properties} & \multicolumn{5}{c|}{RBA Scoring Schemes}  \\ \cline{2-11}

\multicolumn{1}{c||}{} & \multirow{9}{*}{\rotatebox{90}{\textbf{Distance}}} & \multirow{9}{*}{\rotatebox{90}{\textbf{Relevant Feature}}} & \multirow{9}{*}{\rotatebox{90}{\textbf{Irrelevant Features}}} & \multirow{9}{*}{\rotatebox{90}{\textbf{Class}}} & \multirow{9}{*}{\rotatebox{90}{\textbf{Frequency}}}& \multirow{9}{*}{\rotatebox{90}{\textbf{ReliefF}}} & \multirow{9}{*}{\rotatebox{90}{\textbf{SURF}}} & \multirow{9}{*}{\rotatebox{90}{\textbf{SURF*}}} & \multirow{9}{*}{\rotatebox{90}{\textbf{MultiSURF*}}} & \multirow{9}{*}{\rotatebox{90}{\textbf{MultiSURF}}} \\
 
\multicolumn{1}{c||}{}& & & & & & & & & & \\ 
\multicolumn{1}{c||}{}& & & & & & & & & &  \\ 
\multicolumn{1}{c||}{}& & & & & & & & & &  \\ 
\multicolumn{1}{c||}{}& & & & & & & & & &  \\ 
\multicolumn{1}{c||}{}& & & & & & & & & &  \\ 
\multicolumn{1}{c||}{}& & & & & & & & & &  \\ 
\multicolumn{1}{c||}{}& & & & & & & & & &  \\ 
\multicolumn{1}{c||}{Instance Type} & & & & & & & & & &  \\  \hline

$Target$ ($R_{i}$) & - & 0 & . . . & 0 & - & - & -& - & - & - \\  \hline \hline
\rowcolor{gray!30}$Hit_{1}$ & Near & 0 & . . . & 0 & High & & &  &  &  \\  \hline
\rowcolor{gray!30}$Miss_{1}$ & Near & 0 & . . . & 1 & Medium &  & &  &  &   \\  \hline
\rowcolor{gray!30}$Hit_{2}$ & Far & 0 & . . . & 0 & Medium & & & & $-1$ &   \\  \hline
\rowcolor{gray!30}$Miss_{2}$ & Far & 0 & . . . & 1 & Low &  & & & $+1$ &  \\  \hline
$Hit_{3}$ & Near & 1 & . . . & 0 & Low & $- 1$ & $- 1$ & $- 1$ & $- 1$ & $- 1$  \\  \hline
$Miss_{3}$ & Near & 1 & . . . & 1 & Medium & $+1$& $+1$ & $+1$ & $+1$ & $+1$ \\  \hline
$Hit_{4}$ & Far & 1 & . . . & 0 & Medium & & & $+1$ &  &   \\  \hline
$Miss_{4}$ & Far & 1 & . . . & 1 & High & & & $- 1$ &  &   \\  \hline \hline

\multicolumn{1}{|c||}{Total Score} & \multicolumn{5}{c||}{} & \multicolumn{1}{c}{+1} & \multicolumn{1}{|c}{+1} & \multicolumn{1}{|c}{0} & \multicolumn{1}{|c}{0} & \multicolumn{1}{|c|}{+1} \\  \hline

\end{tabular}
} 
\end{table*}

With this in mind, Table \ref{tab:main} lays out the scoring scheme for each algorithm in the right hand columns. First we look at ReliefF, which incidentally has the same scoring scheme as SURF and MultiSURF in the context of this table. Based on the scoring scheme, the only instance types that will contribute to a feature score update are neighbors (i.e. near) with a different feature value than the target ($Hit_{3}$ and $Miss_{3}$), where hits receive $-1$ and misses receive $+1$ every score update. Therefore over all training instance updates, what total score is expected?  For simplicity lets assign our estimated frequencies a numerical value (i.e. high = 3, medium = 2, and low = 1).  Thus ReliefF scoring should yield a total of +1 for a main effect feature (i.e. medium - low).  Since we expect irrelevant features to have a score of approximately 0, we expect ReliefF, as well as SURF and MultiSURF, to be able to distinguish relevant main effect features from irrelevant ones. However, SURF* adds far scoring such that far instances with different feature states also contribute to feature score updates (i.e. $Hit_{4}$ and $Miss_{4}$).  If we add up the frequencies we get an estimated feature score of zero (i.e. medium + medium - high - low). Again assuming that irrelevant features will have a score of approximately 0, it makes sense that SURF* is having difficulty separating relevant from irrelevant features.  The same issue occurs in MultiSURF* that similarly adopts far scoring.  However, here `far' instances with the \emph{same} feature values contribute to scoring to save computational time (i.e. $Hit_{2}$ and $Miss_{2}$). If we add up these frequencies we again get an estimated feature score of zero (i.e. medium + low - medium - low). This conceptual illustration explains our findings and we expect this trend to hold for continuous features and endpoints. Notably, it is possible that this far scoring performance loss may be less apparent when the number of irrelevant features becomes very large. When this happens, relevant features have a much smaller influence on the distance between instances. Regardless, we still always expect ReliefF, SURF, and MultiSURF to perform as good or better than SURF* or MultiSURF* with respect to detecting main effects. 

Next, we will examine a specific example of a simple main effect datasets given in Table \ref{tab:Maindata}. This dataset has the same number of features and instances as given in Table \ref{tab:Epi}. However in this dataset, $A_{1}$ has a strong main effect relevant to class, while $A_{2}$ and $A_{3}$ are irrelevant.  Table \ref{tab:Main2} breaks down nearest and farthest hits and misses in the same way as Table \ref{tab:Epi2}. Table \ref{tab:Mainsum2} presents the summary of the algorithm score contributions in the same way as Table \ref{tab:Episum2}. As demonstrated by the score totals in this example problem, the failure of `far' scoring schemes in SURF* and MultiSURF* is the result of farthest miss (different value) or farthest hit (same value) negative score contributions, respectively. While not explicitly tested in this study we predict that the SWRF* \citep{stokes2012application} algorithm (a related RBA), would similarly suffer from main effect performance loss due to it's adoption of `far' scoring, but this should be explicitly tested in future work. 

\begin{table*}[h!]
 \centering
\caption{Simple example dataset with a relevant main effect in $A_{1}$. Features $A_{2}$ and $A_{3}$ are irrelevant. }
\label{tab:Maindata}       
{\small 
\begin{tabular}{|c||ccc|c|}
\hline
\multicolumn{1}{|c||}{Instances} & \multicolumn{1}{c}{$A_{1}$} & \multicolumn{1}{c}{$A_{2}$} & \multicolumn{1}{c|}{$A_{3}$} & \multicolumn{1}{c|}{$C$}   \\ \hline
$R_{1}$ & 1 & 0 & 1 & 1     \\
$R_{2}$ & 1 & 1 & 0 & 1     \\
$R_{3}$ & 1 & 0 & 1 & 1     \\
$R_{4}$ & 1 & 1 & 0 & 1     \\
$R_{5}$ & 0 & 0 & 1 & 0   \\
$R_{6}$ & 0 & 1 & 0 & 0    \\
$R_{7}$ & 0 & 0 & 1 & 0    \\
$R_{8}$ & 0 & 1 & 0 & 0   \\
\hline
\end{tabular}
}
\end{table*}

\begin{table*}[h!]
 \centering
\caption{Breakdown of near/far, hit/miss, and features with different or same values for each target instance. Illustrates where feature scores come from for the main effect dataset in Table \ref{tab:Maindata}.}
\label{tab:Main2}       

{\tiny 
\begin{tabular}{|c|c|c|c|c|c|c|}
\cline{2-7}
\multicolumn{1}{c|}{} & \multicolumn{2}{c|}{\cellcolor{gray!30}Far (Same Value)}& \multicolumn{2}{c|}{Near (Different Value)} & \multicolumn{2}{c|}{Far (Different Value)} \\ \hline
\multicolumn{1}{|c|}{Target} & \multicolumn{1}{c|}{\cellcolor{gray!30}Hit} & \multicolumn{1}{c|}{\cellcolor{gray!30}Miss}& \multicolumn{1}{c|}{Hit} & \multicolumn{1}{c|}{Miss} & \multicolumn{1}{c|}{Hit} & \multicolumn{1}{c|}{Miss}  \\ \hline

$R_{1}$ &\cellcolor{gray!30}$R_{2}(A_{1})$,$R_{4}(A_{1})$&\cellcolor{gray!30}None&None& $R_{5}(A_{1})$,$R_{7}(A_{1})$ & $R_{2}(A_{2},A_{3})$,$R_{4}(A_{2},A_{3})$ & $R_{8}(A_{1},A_{2},A_{3})$,$R_{6}(A_{1},A_{2},A_{3})$  \\
$R_{2}$ &\cellcolor{gray!30}$R_{1}(A_{1})$,$R_{3}(A_{1})$&\cellcolor{gray!30}None&None& $R_{6}(A_{1})$,$R_{8}(A_{1})$ & $R_{1}(A_{2},A_{3})$,$R_{3}(A_{2},A_{3})$ & $R_{5}(A_{1},A_{2},A_{3})$,$R_{7}(A_{1},A_{2},A_{3})$  \\
$R_{3}$ &\cellcolor{gray!30}$R_{2}(A_{1})$,$R_{4}(A_{1})$&\cellcolor{gray!30}None&None& $R_{5}(A_{1})$,$R_{7}(A_{1})$ & $R_{2}(A_{2},A_{3})$,$R_{4}(A_{2},A_{3})$ & $R_{8}(A_{1},A_{2},A_{3})$,$R_{6}(A_{1},A_{2},A_{3})$  \\
$R_{4}$ &\cellcolor{gray!30}$R_{1}(A_{1})$,$R_{3}(A_{1})$&\cellcolor{gray!30}None&None& $R_{6}(A_{1})$,$R_{8}(A_{1})$ & $R_{1}(A_{2},A_{3})$,$R_{3}(A_{2},A_{3})$ & $R_{5}(A_{1},A_{2},A_{3})$,$R_{7}(A_{1},A_{2},A_{3})$  \\
$R_{5}$ &\cellcolor{gray!30}$R_{6}(A_{1})$,$R_{8}(A_{1})$&\cellcolor{gray!30}None&None& $R_{1}(A_{1})$,$R_{3}(A_{1})$ & $R_{6}(A_{2},A_{3})$,$R_{8}(A_{2},A_{3})$ & $R_{2}(A_{1},A_{2},A_{3})$,$R_{4}(A_{1},A_{2},A_{3})$  \\
$R_{6}$ &\cellcolor{gray!30}$R_{1}(A_{1})$,$R_{3}(A_{1})$&\cellcolor{gray!30}None&None& $R_{2}(A_{1})$,$R_{4}(A_{1})$ & $R_{5}(A_{2},A_{3})$,$R_{7}(A_{2},A_{3})$ & $R_{1}(A_{1},A_{2},A_{3})$,$R_{3}(A_{1},A_{2},A_{3})$  \\
$R_{7}$ &\cellcolor{gray!30}$R_{6}(A_{1})$,$R_{8}(A_{1})$&\cellcolor{gray!30}None&None& $R_{1}(A_{1})$,$R_{3}(A_{1})$ & $R_{6}(A_{2},A_{3})$,$R_{8}(A_{2},A_{3})$ & $R_{2}(A_{1},A_{2},A_{3})$,$R_{4}(A_{1},A_{2},A_{3})$  \\
$R_{8}$ &\cellcolor{gray!30}$R_{5}(A_{1})$,$R_{7}(A_{1})$&\cellcolor{gray!30}None&None& $R_{2}(A_{1})$,$R_{4}(A_{1})$ & $R_{5}(A_{2},A_{3})$,$R_{7}(A_{2},A_{3})$ & $R_{1}(A_{1},A_{2},A_{3})$,$R_{3}(A_{1},A_{2},A_{3})$  \\

\hline
\end{tabular}
}
\end{table*}

\begin{table*}[h!]
 \centering
\caption{Summary of score contributions for main effect dataset in Table \ref{tab:Maindata} yielding ReliefF, SURF, SURF*, MultiSURF*, and MultiSURF scores. Illustrates why `far' scoring hinders main effect detection.}
\label{tab:Mainsum2}       
{\small 
\begin{tabular}{c||c|c|c||c|c|c|c|c|}
\cline{2-9}
\multicolumn{1}{c||}{} & \multicolumn{3}{c||}{Features} & \multicolumn{5}{c|}{RBA Scoring Schemes}  \\ \cline{2-9}
& & & & \multirow{6}{*}{\rotatebox{90}{ReliefF}} & \multirow{6}{*}{\rotatebox{90}{SURF}} & \multirow{6}{*}{\rotatebox{90}{SURF*}} & \multirow{6}{*}{\rotatebox{90}{MultiSURF*}} & \multirow{6}{*}{\rotatebox{90}{MultiSURF}}  \\ 
& & & & & & & &   \\ 
& & & & & & & &   \\ 
& & & & & & & &   \\ 
& & & & & & & &   \\ 
& $A_{1}$&$A_{2}$ & $A_{3}$& & & & &  \\  \hline \hline

\rowcolor{gray!30}Farthest Hit (same value)& 8 & 0 & 0 && && - 1 &   \\
\rowcolor{gray!30}Farthest Miss (same value)& 0 & 0 & 0 && && +1 &\\ \hline
Nearest Hit (different value)& 0 & 0 & 0 & - 1  &- 1 &- 1&- 1&- 1   \\
Nearest Miss (different value)& 8 & 0 & 0 & +1 &+1 &+1&+1&+1  \\ \hline 
Farthest Hit (different value)& 0 & 8 & 8 & &   &+1 &&   \\
Farthest Miss (different value)& 8 & 8 & 8 & &   &- 1&& \\ \hline 
\multicolumn{9}{c}{}\\ \cline{1-4}

\multicolumn{1}{|c||}{ReliefF Score Total} & \multicolumn{1}{c|}{\cellcolor{green!30}8} & \multicolumn{1}{c|}{0} & \multicolumn{1}{c||}{0} & \multicolumn{5}{c}{} \\ \cline{1-4}
\multicolumn{1}{|c||}{SURF Score Total } & \multicolumn{1}{c|}{\cellcolor{green!30}8} & \multicolumn{1}{c|}{0} & \multicolumn{1}{c||}{0} & \multicolumn{5}{c}{} \\  \cline{1-4}
\multicolumn{1}{|c||}{SURF* Score Total } & \multicolumn{1}{c|}{0} & \multicolumn{1}{c|}{0} & \multicolumn{1}{c||}{0} & \multicolumn{5}{c}{} \\  \cline{1-4}
\multicolumn{1}{|c||}{MultiSURF* Score Total } & \multicolumn{1}{c|}{0} & \multicolumn{1}{c|}{0} & \multicolumn{1}{c||}{0} & \multicolumn{5}{c}{} \\ \cline{1-4}
\multicolumn{1}{|c||}{MultiSURF Score Total} & \multicolumn{1}{c|}{\cellcolor{green!30}8} & \multicolumn{1}{c|}{0} & \multicolumn{1}{c||}{0} & \multicolumn{5}{c}{} \\ 

\cline{1-4}
\end{tabular}
}
\end{table*}

\section{Conclusions and Future Study} \label{conclusions}
This work has made a number of contributions with respect to feature selection and RBAs. Specifically we have (1) introduced ReBATE as an open source, user-friendly, and data-type flexible software package for applying a variety of RBAs, (2) designed and applied the most expansive simulation study of RBAs to date, (3) implemented and evaluated strategies to extend RBAs previously only designed for clean data with discrete features and endpoints to a variety of different dataset types, (4) compared the performance of select RBAs to other established feature selection algorithms, (5) identified known or suspected reasons for differences in algorithm performance to guide ongoing RBA development, and (6) introduced and evaluated the MultiSURF algorithm, identifying it to have significant, reliable power in the greatest diversity of dataset configurations. 

The results of this study support the following conclusions; (1) existing popular feature selection methods (i.e. chi square, ANOVA, mutual information, ExtraTrees, and RFE ExtraTrees) fail to perform, or perform as competitively on a variety of generalizable problem types (2) RBAs are proficient at detecting 2-way epistatic interactions, but MultiSURF* in particular performs best in this regard, (3) `far' scoring in RBAs (i.e. SURF* and MultiSURF*) improves the detection of 2-way epistatic interactions (4) `far' scoring deteriorates or even eliminates the ability of SURF* and MultiSURF* to detect simple main effects (5) RBAs function in the presence of patterns of heterogeneous association, (6) the number of neighbors used in RBA scoring is a critical and problem dependent factor with respect to algorithm success, (7) only RBAs that use fewer neighbors in scoring can detect 3-way interactions (i.e. ReliefF 10 NN, and MultiSURF), (8) all implemented data-type expansions in ReBATE were successful, however performance losses are still observed in datasets with a mix of discrete and continuous features, (9) MultiSURF and ReliefF are the only examined methods that can detect all of the following; main effects, heterogeneity, and 2 or 3-way interactions, (10) the main drawback of ReliefF is that the user has to specify a $k$ parameter which our results indicate can dramatically impact success depending on the noise (e.g. heritability), number of training instances, size of the feature space, the heterogeneity ratio, the dimensionality of the interaction, and/or the amount of missing data, (11) the main drawback of MultiSURF is that it failed to detect 4-way and 5-way interactions, and (12) MultiSURF is the most generally flexible and successful method as well as being easier to apply successfully in contrast with ReliefF.

Furthermore, while the asymptotic time complexity of core RBAs is $\mathcal{O}(n^2\cdot a)$, the complete time complexity of MultiSURF is slightly less than that for MultiSURF* which is the best RBA for detecting 2-way epistasis. MultiSURF also appears to scale competitively in feature spaces of increasing size. In Table \ref{tab:rec} we translate the findings of this study into a set of general recommendations for the application of RBA feature selection. 

\begin{table*}[h]
 \centering
\caption{General Recommendations for RBA Application}
\label{tab:rec}       
{\footnotesize 
\begin{tabular}{|l|}
\hline

\rowcolor{gray!30}$\bullet$ RBAs in ReBATE can be applied to feature selection in any bioinformatics problem with \\
\rowcolor{gray!30}  potentially predictive features and a target outcome variable.\\

$\bullet$ Expect RBAs to impact downstream modeling by improving predictive accuracy and model\\
 simplicity/interpretability, particularly when complex patterns of association are present in data.  \\
\rowcolor{gray!30}$\bullet$ If you want to detect feature interactions, but an exhaustive examination of all feature combinations\\
\rowcolor{gray!30} is not feasible, an RBA offers a good alternative. \\
 
$\bullet$ RBAs do not remove redundant/correlated features. It may be most efficient to remove highly  \\
correlated features prior to RBA analysis based on the aims of an analysis. \\

\rowcolor{gray!30}$\bullet$ No single `best' feature selection algorithm for all problems. ReliefF is still the most commonly\\
\rowcolor{gray!30} applied RBA. For simplicity or when computational resources are limited, select MultiSURF instead of \\
\rowcolor{gray!30}ReliefF since it performs well on the greatest diversity of problems and it has no run parameters to be \\
\rowcolor{gray!30} optimized.\\
$\bullet$ If possible, consider running multiple feature selection algorithms with different strengths, e.g. \\
MultiSURF* for 2-way epistasis, ReliefF for higher order interactions, and a myopic method for  \\
main effects, and pass the top non-overlapping set of features returned from each onto modeling. \\

\rowcolor{gray!30}$\bullet$ Roughly, if you have more than 10,000 features, an iterative Relief approach such as TuRF \citep{moore2007tuning}   \\
\rowcolor{gray!30}should be used with a core RBA to overcome the expected performance degradation in large feature\\
\rowcolor{gray!30} spaces. Iterative RBAs are expected to improve RBA performance even in smaller feature spaces.\\
\rowcolor{gray!30} Apply whenever computationally feasible. \\

$\bullet$ RBA's are well suited to detecting features involved in heterogeneous patterns of association such as \\
genetic heterogeneity. This is important for disease subgroup identification.  \\

\rowcolor{gray!30}$\bullet$ For data with both discrete and continuous features, be aware that continuous feature scores will be \\
\rowcolor{gray!30} underestimated in mixed data. We have implemented our own custom ramp function inspired by \\
\rowcolor{gray!30}\citet{hong1997use} into ReBATE to compensate for this observation. \\

$\bullet$ If only interested in detecting main effects, RBAs may be outperformed by traditional myopic \\
methods, particularly in very large feature spaces.\\

\hline
\end{tabular}
}
\end{table*}

Beyond MultiSURF, the results of this study strongly suggest that the RBA concept can be improved further.  As such we have a number of suggested targets for future research; (1) explicitly compare alternate or novel strategies for handling missing data, regression, and mixed feature types, (2) alternative strategies to improve the performance of RBAs in detecting higher dimensional interactions (e.g. 4-way interactions and beyond), (3) given what we have learned in this study, evaluate the impact of instance pair distance weighted scoring similar to strategies proposed in SWRF* and some other core RBAs \citep{robnik1997adaptation,draper2003iterative,sun2006iterative,sun2007iterative,stokes2012application}, (4) consider alternate strategies to adapt neighbor selection in different problems to maximize performance in detecting both main and interaction effects, as pioneered in \citet{mckinney2013reliefseq}, and (5) integrate MultiSURF and other promising RBAs with iterative RBAs to determine the best combination(s) for scaling up to very large feature spaces and identify the practical feature space size boundaries at which we can expect RBA performance in detecting main or interaction effects to become unreliable. In future work we will also seek to expand the diversity of our simulation studies even further and we recommend that other feature selection investigations adopt similar approaches to evaluating and comparing methods.

\section*{Acknowledgements}
Special thanks to Drs. William La Cava and Brian Cole for their constructive feedback and to Weixuan Fu for his contributions to the ReBATE code.  We also thank the Penn Medicine Academic Computing Services for the use of their computing resources. This work was supported by National Institutes of Health grants AI116794, DK112217, ES013508, EY022300, HL134015, LM009012, LM010098, LM011360, TR001263, and the Warren Center for Network and Data Science.





\bibliographystyle{elsarticle-harv}

\bibliography{main}

\end{document}